\documentclass{article}
\usepackage{iclr2027_conference,times}

\usepackage{amsmath,amsfonts,bm}

\def\eqref#1{equation~\ref{#1}}

\def\1{\bm{1}}

\DeclareMathAlphabet{\mathsfit}{\encodingdefault}{\sfdefault}{m}{sl}
\SetMathAlphabet{\mathsfit}{bold}{\encodingdefault}{\sfdefault}{bx}{n}

\usepackage{graphicx}
\usepackage{booktabs}
\usepackage{multirow}
\usepackage{subcaption}
\usepackage{float}
\usepackage{placeins}
\usepackage{xspace}
\usepackage{colortbl}
\usepackage{hyperref}
\usepackage{url}
\usepackage{wrapfig}

\DeclareUnicodeCharacter{2212}{\ensuremath{-}}

\newcommand{\ourabbr}{IMU4D\xspace}
\newcommand{\eg}{\emph{e.g.}\xspace}
\newcommand{\ie}{\emph{i.e.}\xspace}

\definecolor{best_color}{RGB}{253,155,154}
\definecolor{second_color}{RGB}{254,205,158}
\definecolor{third_color}{RGB}{255,255,163}

\title{IMU4D: 4D Human-Object Understanding from Wearable IMUs}

\author{Hao-Yu Hsu$^*$ \quad Tianhang Cheng$^*$ \quad Jing Wen \quad Alexander G. Schwing \quad Shenlong Wang \\
University of Illinois Urbana-Champaign \\
{\small\texttt{\{haoyuyh3,tcheng12,jw116,aschwing,shenlong\}@illinois.edu}} \quad {\normalfont\small $^*$Equal contribution}}

\iclrfinalcopy
\begin{document}

\maketitle
\vspace{-1\baselineskip}

\begin{abstract}
Understanding human activities and the objects they interact with typically relies on visual perception, yet cameras pose persistent challenges in privacy, safety, energy efficiency, and scalability. We explore an alternative: 4D perception without vision. Its goal is to recover human motion and the objects the person interacts with purely from everyday wearable sensors. For this we introduce \ourabbr, a framework that repurposes large language models for non-visual spatiotemporal understanding of human-object dynamics. \ourabbr uses data from a few inertial sensors from earbuds, watches, or smartphones and predicts detailed 4D human motion together with surrounding objects the person interacts with. Experiments across diverse human motion datasets and human-object datasets show that \ourabbr yields more coherent and temporally stable results than SoTA cascaded pipelines, suggesting wearable motion sensors alone can support 4D understanding of human–object interactions. 
\end{abstract}

\section{Introduction}

\begin{figure}[H]
    \centering
    \includegraphics[width=1.0\linewidth]{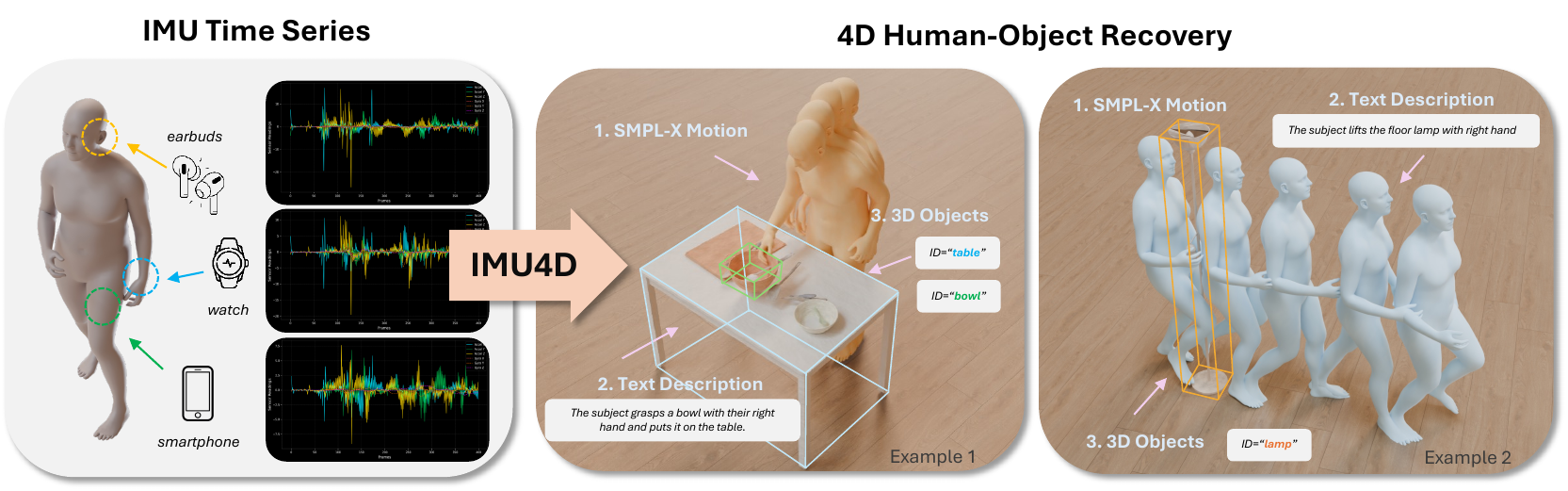}
    \caption{\small
    \textbf{\ourabbr} jointly infers 4D human–object dynamics from only a few everyday wearable IMUs. Given sparse inertial signals from devices such as earbuds, smartphones, or watches, our method predicts SMPL-X body motion, textual activity descriptions, and the category and 6-DoF pose of the 3D objects the person interacts with.}
    \label{fig:teaser}
\end{figure}

Recent advances in multimodal foundation models have expanded how AI perceives and assists in daily life. Yet their understanding of the physical world remains limited: most rely on vision-centric sensing such as cameras or LiDAR, which, while informative, introduces persistent challenges w.r.t.\ privacy, safety, energy efficiency, and scalability. As AI systems become more integrated into daily life, there is an increasing demand for non-visual, privacy-preserving, and low-power modalities that can still capture the richness of human–environment interaction.

Imagine an assistant that can recall how many times you drank water today, where you placed your keys, or whether you’ve been sitting too long—without recording a single image. Such a system would extend AI’s utility to personal health, memory, and accessibility. This motivates a research question: \textit{How much can everyday motion sensors embedded in earbuds, watches, and smartphones reveal about our activities and environments?}

Existing approaches (Tab.~\ref{tab:task_compare}) address only part of this problem. IMU-based motion estimators~\citep{xu2024mobileposer} recover body motion from commodity devices, but output neither activity semantics nor objects. Egocentric understanding methods~\citep{hong2025egolm} jointly predict motion and activity descriptions, but rely on egocentric video, reintroducing the privacy cost of cameras. Motion-to-text~\citep{zhu2025motiongpt3} and motion-to-object methods~\citep{ye2022scene} assume clean, ground-truth motion as input. Chaining these models is a natural baseline, yet errors in the estimated motion propagate to every downstream output, and the separately trained stages cannot benefit from one another.

To address this gap, we propose \textbf{\ourabbr} (Fig.~\ref{fig:teaser}), a framework for vision-free 4D perception that infers detailed human motion and the 3D objects involved in the interaction purely from inertial signals. The key insight behind \ourabbr: \textit{human motion and the objects involved in an interaction are inherently coupled}—inertial observations constrain how a person moves, while learned motion–object associations provide priors over the objects and configurations compatible with that motion. We exploit these complementary sources of information to jointly infer human motion, activity descriptions, and interacting objects.

Recent multimodal foundation models~\citep{deng2025bagel,wang2024emu3} have demonstrated that such coupled structure across modalities can be effectively captured by leveraging rich pre-trained representations from large language models (LLMs) and extending them to new modalities through unified architectures. Building on this intuition, \ourabbr repurposes an LLM as a spatiotemporal reasoner, jointly modeling the distribution of sequential inertial signals, human motions, 3D objects the person interacts with, and semantic activities. %
To achieve this, we map each modality (IMU, motion, and 3D objects) into the embedding space of the LLM and formulate a unified sequential prediction problem, where the model first recovers human motion, then describes the activity in text, and finally infers the objects involved and their poses over time. The model is jointly finetuned with a multi-task loss in a two-stage training process. This unified formulation allows the model to infer coherent, context-aware 4D representations from as few as three inertial sensors—substantially improving temporal and spatial consistency compared with cascaded pipelines (IMU $\rightarrow$ motion $\rightarrow$ 3D object). 
Through the compositional reasoning capacity of LLMs, \ourabbr bridges physical dynamics with spatial and semantic context, enabling a joint understanding of human-object interaction without relying on visual input.

Experiments on diverse datasets and real-world IMU recordings show that \ourabbr achieves more coherent and temporally consistent recovery than prior approaches. Our results demonstrate that ubiquitous wearable sensors alone can support privacy-preserving 4D understanding of human activities and the surrounding objects involved in the interaction—paving the way for embodied AI that perceives the world through motion rather than sight.

Our contributions are threefold:
\begin{itemize}
\item We introduce the first framework that recovers 4D human motion, textual descriptions, and the categories and poses of objects the person interacts with from as few as three IMU sensors by repurposing LLMs for cross-modal structural reasoning.
 \item We formulate IMU-based 4D perception as a unified sequence prediction problem that jointly decodes human motion, text, and a variable number of objects together with their per-frame trajectories.
\item Experiments show that our joint reasoning framework outperforms pipelined SoTA methods and enables recognition and localization of interacted objects without any visual input.
\end{itemize}

\section{Related Work}

\begin{table}[t]
\caption{\small\textbf{Motion- and object-understanding task comparison.}}
\centering
\resizebox{0.9\linewidth}{!}{
\setlength{\tabcolsep}{5pt}
\begin{tabular}{l|c|ccc}
\toprule
\multirow{2}{*}{Task} & \multirow{2}{*}{Input Modality} & \multicolumn{3}{c}{Output Modality} \\
\cmidrule(lr){3-5}
 & & Motion & Text & 3D Object \\
\midrule
Ego-Understand~\citep{hong2025egolm,yi2025estimating} & Ego Video + Sensors & \checkmark & \checkmark & \\
Motion-to-Text~\citep{zhu2025motiongpt3,jiang2023motiongpt} & GT Motion & \checkmark & \checkmark & \\
Motion-to-Object~\citep{ye2022scene,nie2022pose2room} & GT Motion & & & \checkmark \\
IMU-to-Motion~\citep{xu2024mobileposer,mollyn2023imuposer} & Raw IMU & \checkmark & & \\
\textbf{\ourabbr (Ours)} & Raw IMU & \checkmark & \checkmark & \checkmark \\
\bottomrule
\end{tabular}
}
\label{tab:task_compare}
\end{table}

\noindent{\textbf{IMU-Based Human Motion Estimation.}}~
Early works~\citep{sony2025mocopi, von2017sparse, huang2018deep, yi2021transpose, dittadi2021full, yi2022physical, jiang2022avatarposer, jiang2022transformer, zuo2024loose, yi2024physical, van2024diffusionposer, xiao2024fast, yi2025improving, zuo2025transformer} explored IMU-based motion capture as a cost-effective alternative to optical systems~\citep{moeslund2001survey}, but typically rely on specialized body-worn sensors~\citep{schepers2018xsens}. Recent works~\citep{mollyn2023imuposer, xu2024mobileposer, zhang2025baroposer, boscolocamiletto2026towards} instead leverage commodity devices such as smartphones, smartwatches, and wireless earbuds, enabling more accessible motion sensing but remaining largely task-specific and lacking scene awareness. Other approaches incorporate egocentric video~\citep{hong2025egolm, zhang2024masked, yi2025estimating, lee2024mocap, patel2025uniegomotion} to provide richer contextual cues, at the cost of additional sensing and privacy concerns. In contrast, \ourabbr jointly reasons about human motion, interacting 3D objects, and semantics using only sparse IMU signals from everyday devices, without visual input.

\noindent{\textbf{Motion-Object Synthesis.}}
Human motion and interacting objects are strongly coupled both semantically and geometrically. Existing human-object interaction datasets~\citep{bhatnagar2022behave, li2023omomo, zhang2024hoi, xu2025interact, lu2025humoto, kim2025parahome} enable generating motion conditioned on objects~\citep{hassan2021stochastic, li2023omomo, zhao2023synthesizing, kulkarni2024nifty, jiang2024scaling, ghosh2023imos} and inferring object configurations from motion~\citep{nie2022pose2room, ye2022scene, yi2022human, yi2023mime, li2024physics}. Recent methods further jointly model humans and objects~\citep{petrov2025tridi, xu2025interact, xu2024interdreamer,li2026pami,eskandar2026cosmi}, with physics~\citep{merel2020catch, lee2023locomotion, hassan2023synthesizing, braun2024physically} or language~\citep{wang2024move, jiang2024autonomous, li2024controllable, xu2024interdreamer} providing additional constraints. However, these methods typically assume clean human poses from motion capture or simulation. In contrast, \ourabbr directly reasons about human motion and interacting 3D objects from sparse, raw inertial measurements.

\noindent{\textbf{Large Multimodal Foundation Models.}}
Large multimodal foundation models unify diverse modalities into shared representations for joint understanding and generation~\citep{liu2023llava, xie2024show, deng2025bagel}, achieving success across vision~\citep{liu2023llava, xie2024show, deng2025bagel, chen2025janus, wang2024emu3, wu2024vila, zhou2024transfusion, liu2024world, qu2025tokenflow, wu2024next, ge2024seed, bai2025qwen2, team2024chameleon, chen2024expanding}, audio~\citep{ghosh-etal-2024-gama, zhang2023speechgpt}, and robotics~\citep{kim2024openvla, intelligence2504pi0}. Recent works have extended this paradigm to inertial signals~\citep{hong2025egolm, zhang2026sensorlm, xu2026inertia} and human motion~\citep{zhu2025motiongpt3, zhang2024motiongpt, chen2024motionllm, zhang2024large, wang2024scaling, lu2025scamo, li2025genmo, shi2025genm, fan2025go, guo2025snapmogen}. However, their use for ambient sensing and human-object perception remains largely underexplored. \ourabbr bridges this gap by jointly reasoning about inertial sensing, human motion, and surrounding 3D objects within a unified multimodal framework.

\section{Method}
\label{sec:method}

\begin{figure*}[t]
    \centering
    \includegraphics[width=\linewidth]{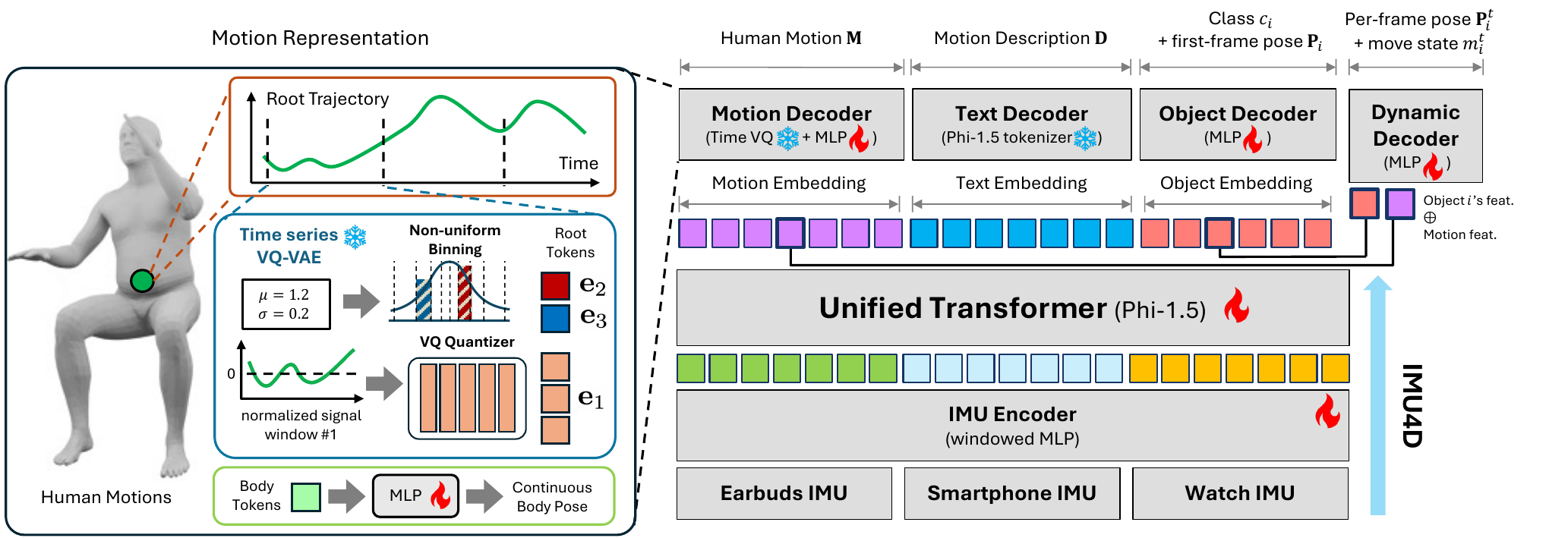}
    \caption{\small
    \textbf{Overview of \ourabbr.} (Right) IMU streams from earbuds, smartphone, and watch are encoded and fed into a unified transformer, which jointly decodes human motions, motion descriptions, and objects. A dynamic decoder then combines each object's feature with the motion features to predict its per-frame pose and moving state. (Left) Motion representation. The root trajectory is split into windows; each window is normalized and VQ-quantized ($\mathbf{e}_1$), while its mean $\mu$ and standard deviation $\sigma$ are quantized by non-uniform binning ($\mathbf{e}_2$, $\mathbf{e}_3$). Body poses are regressed continuously.
    }
    \label{fig:method}
\end{figure*}

\ourabbr casts IMU-based 4D perception as a single sequence-prediction problem over a pretrained LLM. From the IMU signals, it decodes human motion, a text description, and a list of objects with their trajectories (Fig.~\ref{fig:method}). We formalize the task in Sec.~\ref{sec:prob_form}, describe the unified sequence model in Sec.~\ref{sec:arch} and the modality-specific encoders and decoders in Sec.~\ref{sec:tokenize}, and present the training procedure in Sec.~\ref{sec:train}.

\subsection{Problem Formulation}
\label{sec:prob_form}

Formally, we learn a model $f_{\boldsymbol{\theta}}$ that maps IMU signals $\mathbf{S}$ to human motion $\mathbf{M}$, a text description $\mathbf{D}$, and the set of interacted objects $\mathcal{O}$:
\begin{equation}
(\mathbf{M}, \mathbf{D}, \mathcal{O}) = f_{\boldsymbol{\theta}}(\mathbf{S}).
\label{eq:formulation}
\end{equation}
The input $\mathbf{S}\in\mathbb{R}^{T \times N_\mathrm{imu} \times 15}$ contains $T$ frames from $N_\mathrm{imu}\in\{1,\ldots,5\}$ sensors worn at the ears, wrists, and thighs, corresponding to earbuds, watches, and smartphones placed in pockets. Each sensor provides a 3D linear acceleration, a 3D angular velocity, and a $3\times3$ orientation matrix. We assume no magnetometer, as many wearable devices do not provide reliable and accessible magnetometer data, and such readings are often corrupted by environmental magnetic interference.

The {\bf motion} $\mathbf{M}\in\mathbb{R}^{T\times 69}$ consists of the per-frame global translation, global orientation, and 21 SMPL-X~\citep{pavlakos2019expressive} joint rotations in axis-angle form; internally, we use the 6D rotation representation~\citep{zhou2019continuity} for stable training. The {\bf text description} $\mathbf{D}$, also referred to as the caption, is a sequence of text tokens describing the observed activity.
The {\bf objects} form a variable-size set $\mathcal{O}=\{\mathbf{O}_i\}_{i=1}^{N_\text{obj}}$, where $N_\text{obj}$ is predicted by the model. We predict each object's category and pose, but do not reconstruct its shape. Each object $\mathbf{O}_i$ consists of a class $c_i\in\{1,\ldots,C\}$, a first-frame pose $\mathbf{P}_i\in\mathbb{R}^{9}$ (6D orientation and 3D translation), and per-frame poses $\mathbf{P}_i^t$ with moving states $m_i^t\in\{0,1\}$ for $t=1,\ldots,T$, which indicate whether the object stays in place (\eg, a table) or is being carried (\eg, a mug).

All sequences are canonicalized such that, at the first frame, the pelvis is at the origin and faces the $+\mathrm{Z}$ axis. The root translation and the objects' first-frame poses are expressed in this canonical frame, while object trajectories are expressed relative to the body root at each frame.

\subsection{Unified Sequence Modeling}
\label{sec:arch}
We build \ourabbr on Phi-1.5~\citep{li2023textbooks} (1.3B), initialized from the pretrained Show-o~\citep{xie2024show} checkpoint. All modalities are mapped into the LLM's embedding space and placed in a single sequence: the IMU signals, followed by the motion, the caption, and the objects. This order follows how directly each output is constrained by the IMU signals. Motion is observed most directly, so it comes first; the caption summarizes the motion; and objects, the most ambiguous, are inferred last from both (Tab.~\ref{tab:joint_formulations}).

Each output is decoded in the way that suits it. Motion has the same length as the input, so the model decodes it in a single parallel pass rather than frame by frame, which avoids accumulating errors over long sequences. The caption and the object list have variable lengths, so they are generated autoregressively until an end token (\texttt{<EOT>} and \texttt{<EOOBJ>}), letting the model decide how many objects are present.

\subsection{Modality Encoders and Decoders}
\label{sec:tokenize}
We now describe how each modality is mapped into and out of the shared sequence.

\noindent{\textbf{IMU Encoder.}}
Given raw IMU signals of length $T$, we divide them into $T/d_{\mathrm{win}}$ non-overlapping windows ($d_{\mathrm{win}}=4$) and project each window into the transformer's embedding space with an MLP.
Each device is encoded independently, and the resulting embeddings are concatenated to form the full input sequence. Embeddings corresponding to inactive IMU devices are replaced with $\texttt{<MASK>}$ embeddings.

\noindent{\textbf{Motion Decoder.}}
We split human motion into two parts with different statistics (Fig.~\ref{fig:method}, left). The \emph{body pose}, \ie, the 21 local joint rotations, consists of absolute angles within a bounded range, so we regress it directly as continuous 6D rotations. The \emph{root trajectory}, \ie, the per-frame increments of global translation and orientation, has a large dynamic range and is integrated over time, so small regression errors compound into drift. We therefore discretize it with a time-domain tokenizer. Existing time-domain tokenizers either quantize each frame independently~\citep{dwivedi2024tokenhmr}, ignoring temporal coherence, or normalize with dataset-wide statistics~\citep{xiao2025motionstreamer}, which leaves the normalized signals too diverse for a fixed-size codebook. We instead split the root trajectory into $d_\mathrm{win}$-frame windows aligned with the IMU windows, and normalize each window by its own mean $\mu$ and standard deviation $\sigma$. For each channel, a pretrained, frozen time-series VQ-VAE~\citep{talukder2024totem} quantizes the normalized shape into a token $\mathbf{e}_1$, and $\mu$ and $\sigma$ are quantized by non-uniform binning into tokens $\mathbf{e}_2$ and $\mathbf{e}_3$. Per-window normalization makes the shapes compact enough for precise quantization, while $\mathbf{e}_2$ and $\mathbf{e}_3$ restore the scale and offset needed to recover the trajectory. 
The motion segment consists of learnable queries, one per $d_\mathrm{win}$-frame window and aligned with the IMU windows, which attend to the IMU tokens and bidirectionally to each other. Each query predicts the root tokens of its window with classification heads and the body poses of its $d_\mathrm{win}$ frames with an MLP regressor.

\noindent{\textbf{Text Decoder.}}
We use the tokenizer and detokenizer of the pretrained Phi-1.5 model~\citep{li2023textbooks} for text, and extend its vocabulary with special tokens that mark the boundaries of each modality.

\noindent{\textbf{Object and Dynamic Decoders.}}
We decode each object $\mathbf{O}_i$ (Sec.~\ref{sec:prob_form}) in three steps: its class, its first-frame pose, and its per-frame trajectory with moving states. Objects are generated autoregressively after the caption. At each step, the model predicts the class $c_i$ of the next object, or the stop token \texttt{<EOOBJ>}, conditioned on all previous tokens. 
The predicted class is mapped to a learned embedding and fed back as the next input. From the corresponding hidden state, an MLP \emph{object decoder} regresses the object's first-frame pose, which is also embedded and fed back, so that later objects are conditioned on earlier placements.
Finally, a \emph{dynamic decoder} predicts each object's per-frame pose and moving state from its feature and the motion features; the state is predicted per frame because an object can be picked up and later put down. Static objects stay in place, while moving objects follow the body.

\subsection{Training}
\label{sec:train}

\noindent{\textbf{Objective.}}
The model is trained end-to-end with
\begin{equation}
\mathcal{L} = \mathcal{L}_\text{motion} + \mathcal{L}_\text{text} + \mathcal{L}_\text{obj}
\label{eq:loss}
\end{equation}
where $\mathcal{L}_\text{motion}$ combines CE loss on the root tokens of each window with L1 loss on the body poses, $\mathcal{L}_\text{text}$ is the next-token CE loss on the caption, and $\mathcal{L}_\text{obj}$ combines CE loss on object classes, L1 loss on first-frame poses and per-frame trajectories, and BCE loss on moving states. Each term is applied only to the modalities annotated in a given sample. Loss weights and the full formulation are provided in Appendix~\ref{sec:supp_implementation_details}.

\noindent{\textbf{Stage 1: Mixed-dataset pretraining.}}
Starting from pretrained Show-o, we train a single model on a mixture of datasets, including large-scale motion--text data and human--object interaction data. For datasets without IMU recordings, we synthesize IMU signals by attaching virtual sensors to the body following~\citet{yi2025improving}. Since the datasets differ in their annotations, \eg, only the interaction datasets provide objects, each sample is supervised only on the modalities it contains (Eq.~\ref{eq:loss}). This allows object prediction, for which data are scarce, to benefit and build upon representation learned from the abundant motion--text data.

\noindent{\textbf{Stage 2: Task-specific finetuning.}}
Since interaction and real-IMU data make up only a small fraction of the mixture, we then finetune the pretrained model separately on each evaluation dataset to adapt it to that dataset's motion, object, and sensor distributions for better performance.

\begin{table*}[t]
\caption{\small\textbf{Quantitative comparison of IMU-to-Motion on real-world IMU datasets DIP-IMU~\citep{huang2018deep} and IMUPoser~\citep{mollyn2023imuposer}.} 
We omit MTE following MobilePoser~\citep{xu2024mobileposer}. Evaluated on whole sequences using a 3pt setup (ear, wrist, thigh).}
\centering
\resizebox{0.95\linewidth}{!}
{
\setlength{\tabcolsep}{5pt}
\begin{tabular}{@{} l cccc cccc @{}}
\toprule
& \multicolumn{4}{c}{DIP-IMU} & \multicolumn{4}{c}{IMUPoser} \\
\cmidrule(lr){2-5} \cmidrule(lr){6-9}
Method 
& MPJPE $\downarrow$ & PA-MPJPE $\downarrow$ & MPJRE $\downarrow$ & MPJVE $\downarrow$
& MPJPE $\downarrow$ & PA-MPJPE $\downarrow$ & MPJRE $\downarrow$ & MPJVE $\downarrow$ \\
\midrule
IMUPoser & 64.25 & 53.13 & 9.06 & 85.83 & 66.86 & 52.62 & 9.81 & 84.24 \\
MobilePoser & 62.91 & 51.33 & 7.95 & 81.34 & 61.96 & 50.07 & 8.44 & 75.59 \\
Ours & \textbf{47.31} & \textbf{39.97} & \textbf{5.31} & \textbf{58.71} & \textbf{42.45} & \textbf{37.43} & \textbf{5.70} & \textbf{54.71} \\
\bottomrule
\end{tabular}
}
\label{tab:quant_imu2motion}
\end{table*}

\ifdefined\meetingimg\else\newsavebox{\meetingimg}\fi
\begin{figure*}[t]
  \centering
  \sbox{\meetingimg}{\includegraphics[width=0.96\textwidth]{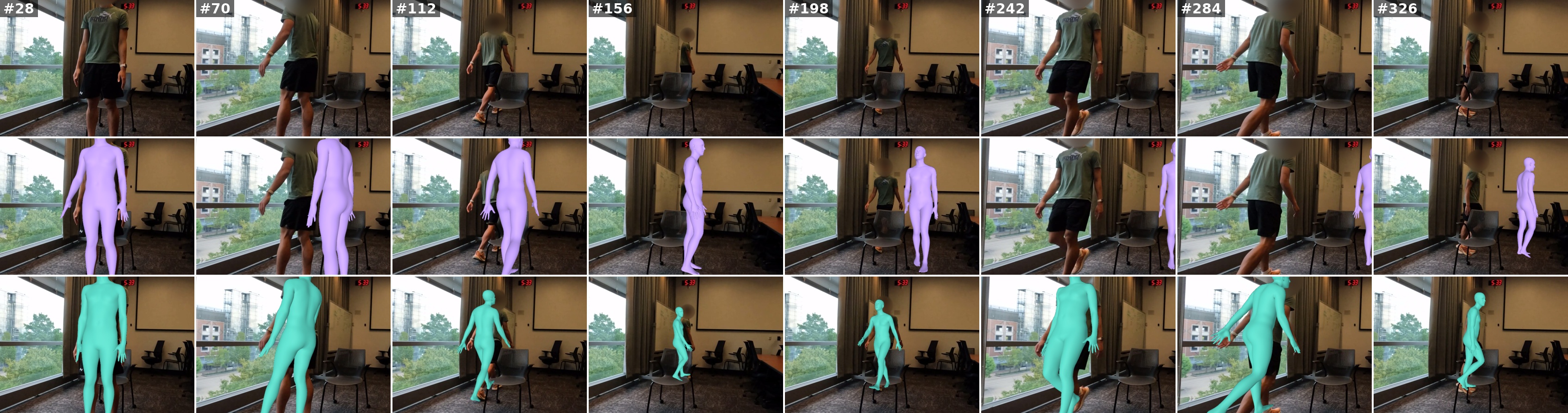}}%
  \begin{minipage}[c]{1em}
    \offinterlineskip\setlength{\parindent}{0pt}%
    \parbox[c][\dimexpr\ht\meetingimg/3\relax][c]{1em}{\centering\rotatebox{90}{\scriptsize RGB}}\par
    \parbox[c][\dimexpr\ht\meetingimg/3\relax][c]{1em}{\centering\rotatebox{90}{\scriptsize MobilePoser}}\par
    \parbox[c][\dimexpr\ht\meetingimg/3\relax][c]{1em}{\centering\rotatebox{90}{\scriptsize\bfseries Ours}}\par
  \end{minipage}\hspace{0.5mm}%
  \raisebox{-0.5\ht\meetingimg}{\usebox{\meetingimg}}
  \vspace{-2mm}
  \caption{\small\textbf{Real-world capture in a meeting room.} From three consumer IMUs, MobilePoser
  drifts away from the subject and leaves the frame, whereas our prediction stays registered to the
  person throughout the sequence. Numbers denote frame indices at 30\,fps; the RGB frames are shown
  only for visualization and are not used as input.}
  \label{fig:meeting_room}
\end{figure*}

\section{Experiments}
\label{sec:exp}
In this section, we start with the implementation details and evaluation setups. Then we evaluate \ourabbr on 4D human motion, text description, and 3D object predictions across multiple datasets. Our experiments are designed to answer three key questions:
(1) How accurately can \ourabbr infer everyday human activities and their surrounding objects?
(2) Does large-scale pre-training on a foundation model provide advantages over task-specific specialized models?
(3) Does joint reasoning outperform a pipelined approach that cascades state-of-the-art modules?

\subsection{Experimental Settings}

\noindent{\textbf{Tasks and Evaluation Datasets.}}
We evaluate three tasks:
(1) \textbf{IMU-to-Motion (I2M)}, which predicts human motion from IMU signals. We use HumanML~\citep{guo2022generating}, DIP-IMU~\citep{huang2018deep}, IMUPoser~\citep{mollyn2023imuposer} as our evaluation datasets, among which DIP-IMU and IMUPoser provide IMU signals captured in \textit{real-world} settings. 
We additionally collect a real-world commodity-device IMU dataset, \textbf{IMU4D-Real}, using earbuds, smartphones, and smartwatches to evaluate performance under practical sensing conditions.
We evaluate under two configurations: a 3-point (3pt) setup with IMUs mounted on one ear, one wrist, and one thigh, and a 5-point (5pt) setup with IMUs mounted on one ear, both wrists, and both thighs. We report results under 3pt setup on DIP-IMU and IMUPoser, and under 5pt setup on HumanML.
(2) \textbf{IMU-to-MotionText (I2MT)}, which jointly predicts motion sequences and text descriptions from IMU inputs. We conduct our evaluation on HumanML~\citep{guo2022generating}, HUMOTO~\citep{lu2025humoto}, OMOMO~\citep{li2023omomo}, and HiPhi~\citep{ji2026hiphi}. Note that HumanML is a superset of AMASS~\citep{mahmood2019amass}—a dataset widely adopted for IMU pretraining in prior work~\citep{xu2024mobileposer, mollyn2023imuposer}. We use a 5pt setup for I2MT.
(3) \textbf{IMU-to-MotionTextObject (I2MTO)}, which requires holistic reasoning over motion, text, and 3D objects in the interaction.  We evaluate two settings: \emph{static object layouts} on HUMOTO~\citep{lu2025humoto}, where we evaluate the first-frame placement of multiple objects per sequence, and \emph{dynamic objects} on OMOMO~\citep{li2023omomo} and HiPhi~\citep{ji2026hiphi}, which provide per-frame object trajectories.
All data are processed at 30\,fps.

\noindent{\textbf{Evaluation Metrics.}}
We evaluate three output modalities: 4D human motion, text description, and 3D objects. For motion, we follow prior work~\citep{xu2024mobileposer, hong2025egolm} and report joint accuracy (MPJPE, PA-MPJPE), mesh accuracy (MPJVE), and trajectory error (MTE) for long-horizon prediction, all in millimeters, and joint rotation error (MPJRE) in degrees. For text, we adopt the linguistic metrics from MotionGPT3~\citep{zhu2025motiongpt3} and report BLEU~\citep{papineni2002bleu}, ROUGE-L~\citep{lin2004rouge}, CIDEr~\citep{vedantam2015cider}, and BERT-Score~\citep{zhang2019bertscore}. For 3D objects, we report 3D bounding box IoU (3D-IoU) and identity-level precision and recall (ID-P, ID-R) of predicted objects. On OMOMO and HiPhi, we further report the recall of moving objects, the F1 score of dynamic objects at an IoU threshold of 0.25 (F1@0.25), and the object mesh velocity error on moving frames. For more details on metrics, please refer to Appendix~\ref{sec:supp_evaluation_details}.

\noindent{\textbf{Baselines.}}
For IMU-to-Motion, we compare against IMUPoser~\citep{mollyn2023imuposer} and MobilePoser~\citep{xu2024mobileposer}, both using the same IMU placement as ours. 
As no prior work jointly predicts motion, text, and 3D objects from IMU signals, we construct a modular baseline, \textit{Mod-I2MTO}, which uses MobilePoser for motion estimation and MotionGPT3~\citep{zhu2025motiongpt3} for motion captioning, followed by a motion-to-object model suited to each setting. For static layouts on HUMOTO, we use SUMMON~\citep{ye2022scene} and Pose2Room~\citep{nie2022pose2room}, which infer object placements from human motion. For dynamic objects on OMOMO and HiPhi, we use Object Pop-up~\citep{petrov2023objpopup} and TriDi~\citep{petrov2025tridi}, which predict per-frame object poses from the human body.
All baselines are retrained on the same training data as ours.

\subsection{Experimental Results}

\noindent{\textbf{IMU-to-Motion.}} As shown in Tab.~\ref{tab:quant_imu2motion}, our model outperforms all baselines on every motion metric on real-world DIP-IMU and IMUPoser datasets, showing better adaptability to real-world conditions. 
We further evaluate on our commodity device dataset IMU4D-Real in Tab.~\ref{tab:real_world_commodity}, and also demonstrate better robustness over the baselines. Qualitative results in Fig.~\ref{fig:meeting_room} show that our method better tracks the ground-truth motion over time, while MobilePoser drifts away from the subject.

\begin{table*}[t]
\caption{\small\textbf{Quantitative comparison of IMU-to-MotionText using 480-frame IMUs.} We use the 5pt setup (ear, both wrists, both thighs).}
\centering
\resizebox{0.95\linewidth}{!}{
\setlength{\tabcolsep}{2pt}
\begin{tabular}{@{} l l ccccc ccccc @{}}
\toprule
\multirow{2}{*}{Benchmark} & \multirow{2}{*}{Method} & \multicolumn{5}{c}{Motion Metrics} & \multicolumn{5}{c}{Text Metrics} \\
\cmidrule(lr){3-7} \cmidrule(lr){8-12}
& & MPJPE $\downarrow$ & PA-MPJPE $\downarrow$ & MPJRE $\downarrow$ & MPJVE $\downarrow$ & MTE $\downarrow$ & ROUGE-L $\uparrow$ & CIDEr $\uparrow$ & BLEU@1 $\uparrow$ & BLEU@4 $\uparrow$ & BERT $\uparrow$ \\
\midrule
\multirow{2}{*}{HumanML} & Mod-I2MTO & 43.70 & 34.59 & 11.32 & 56.05 & 275.39 & 36.80 & 133.15 & 32.55 & 4.85 & 26.50 \\
 & Ours & \textbf{14.69} & \textbf{11.32} & \textbf{3.06} & \textbf{19.31} & \textbf{48.50} & \textbf{43.35} & \textbf{197.96} & \textbf{38.08} & \textbf{11.00} & \textbf{35.17} \\
\midrule
\multirow{2}{*}{HUMOTO} & Mod-I2MTO & 50.08 & 38.77 & 11.59 & 60.38 & 277.02 & 33.21 & 88.27 & 30.17 & 4.43 & 19.43 \\
 & Ours & \textbf{31.41} & \textbf{25.60} & \textbf{7.51} & \textbf{38.16} & \textbf{61.67} & \textbf{34.09} & \textbf{104.85} & \textbf{35.52} & \textbf{5.79} & \textbf{22.87} \\
\midrule
\multirow{2}{*}{OMOMO} & Mod-I2MTO & 43.06 & 33.74 & 9.97 & 59.97 & 235.57 & 74.21 & 537.99 & 72.71 & 50.22 & 53.66 \\
 & Ours & \textbf{31.25} & \textbf{22.16} & \textbf{6.93} & \textbf{44.53} & \textbf{59.23} & \textbf{79.74} & \textbf{619.57} & \textbf{78.14} & \textbf{59.63} & \textbf{63.45} \\
\midrule
\multirow{2}{*}{HiPhi} & Mod-I2MTO & 43.52 & 32.81 & 11.73 & 58.32 & 228.42 & 36.16 & 115.94 & 34.16 & 6.68 & 20.67 \\
 & Ours & \textbf{21.59} & \textbf{15.70} & \textbf{6.96} & \textbf{29.29} & \textbf{52.11} & \textbf{41.66} & \textbf{186.81} & \textbf{41.21} & \textbf{14.16} & \textbf{24.66} \\
\bottomrule
\end{tabular}
}
\label{tab:quant_imu2all}
\end{table*}

\begin{table*}[t]
\caption{\small\textbf{Real-world commodity-device results on IMU4D-Real dataset.} Whole sequences, 3pt setup (left wrist, right thigh, ear). MobilePoser's text is obtained by captioning its motion with MotionGPT3. N/A: no global translation; --: no text output.}
\centering
\resizebox{\linewidth}{!}{
\setlength{\tabcolsep}{5pt}
\begin{tabular}{@{} l ccccc ccccc @{}}
\toprule
& \multicolumn{5}{c}{IMU-to-Motion} & \multicolumn{5}{c}{IMU-to-Text} \\
\cmidrule(lr){2-6}\cmidrule(l){7-11}
Method & MPJPE $\downarrow$ & PA-MPJPE $\downarrow$ & MPJRE $\downarrow$ & MPJVE $\downarrow$ & MTE $\downarrow$
       & ROUGE-L $\uparrow$ & CIDEr $\uparrow$ & BLEU@1 $\uparrow$ & BLEU@4 $\uparrow$ & BERT $\uparrow$ \\
\midrule
IMUPoser & 94.83 & 66.89 & 13.53 & 121.59 & N/A & -- & -- & -- & -- & -- \\
MobilePoser + MGPT3 & 87.46 & 62.77 & 14.46 & 113.94 & 979.77 & 31.5 & 89.8 & 31.2 & 1.0 & 16.3 \\
Ours & \textbf{37.12} & \textbf{31.30} & \textbf{7.46} & \textbf{50.18} & \textbf{302.41} & \textbf{35.5} & \textbf{100.4} & \textbf{31.5} & \textbf{3.4} & \textbf{17.2} \\
\bottomrule
\end{tabular}
}
\label{tab:real_world_commodity}
\end{table*}

\ifdefined\scenecmplabel\else\newcommand{\scenecmplabel}[1]{\makebox[\scenecmpw]{\tiny #1}}\fi
\begin{figure}[t]
  \centering
  \setlength{\tabcolsep}{0.4pt}%
  \ifdefined\scenecmplab\else\newlength{\scenecmplab}\fi\setlength{\scenecmplab}{1.0em}%
  \ifdefined\scenecmpclip\else\newlength{\scenecmpclip}\fi\setlength{\scenecmpclip}{3pt}%
  \ifdefined\scenecmpw\else\newlength{\scenecmpw}\fi%
  \setlength{\scenecmpw}{\dimexpr(\linewidth-\scenecmplab-1\scenecmpclip-16\tabcolsep)/8\relax}%
  \ifdefined\scenecmph\else\newlength{\scenecmph}\fi\setlength{\scenecmph}{0.8333\scenecmpw}%
  \parbox[b][\scenecmph][c]{\scenecmplab}{\centering\rotatebox{90}{\scriptsize\textbf{HUMOTO}}}%
\begin{tabular}[b]{@{}cccccccc@{}}
\scenecmplabel{GT} &
\scenecmplabel{Baseline 1} &
\scenecmplabel{Baseline 2} &
\scenecmplabel{\textbf{Ours}} &
\hspace{\scenecmpclip}\scenecmplabel{GT} &
\scenecmplabel{Baseline 1} &
\scenecmplabel{Baseline 2} &
\scenecmplabel{\textbf{Ours}} \\[-0.6ex]
\includegraphics[width=\scenecmpw]{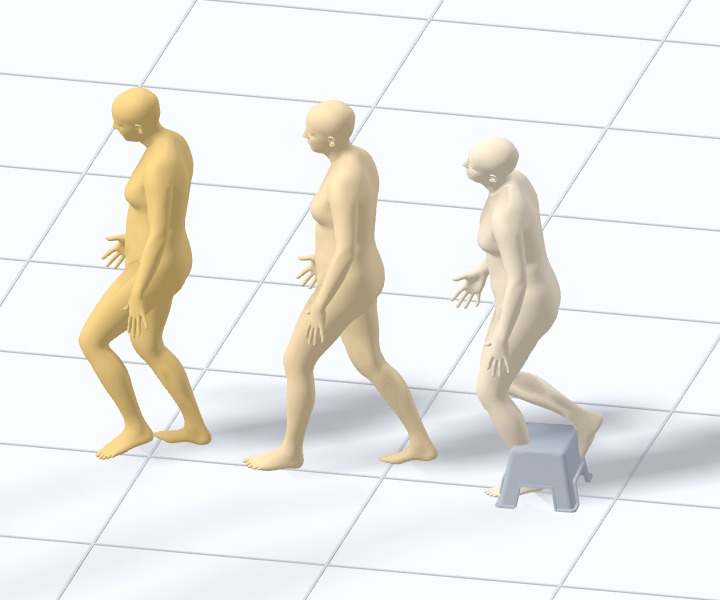} &
\includegraphics[width=\scenecmpw]{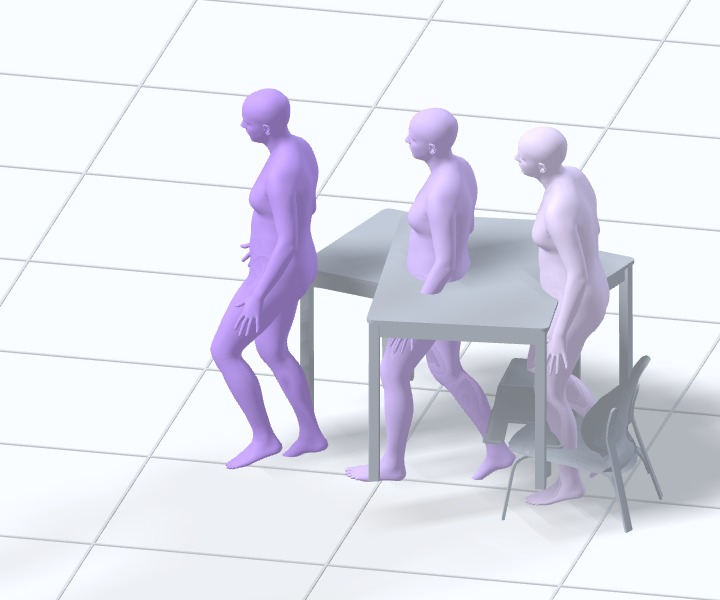} &
\includegraphics[width=\scenecmpw]{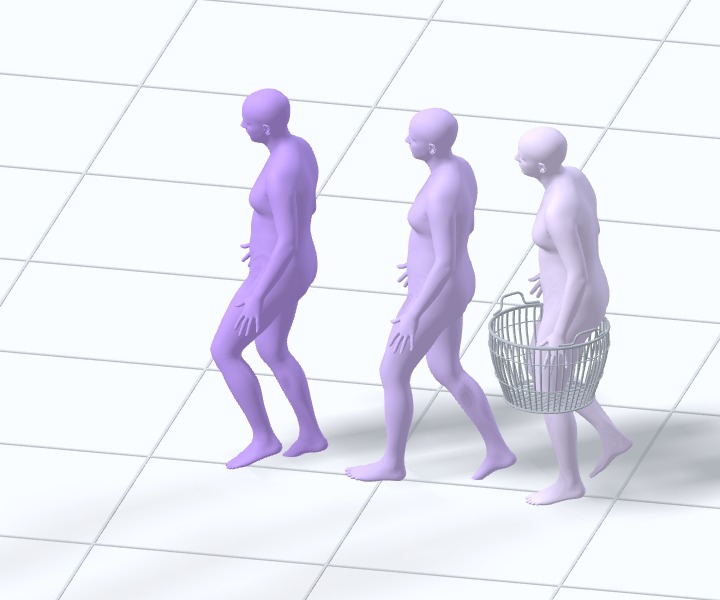} &
\includegraphics[width=\scenecmpw]{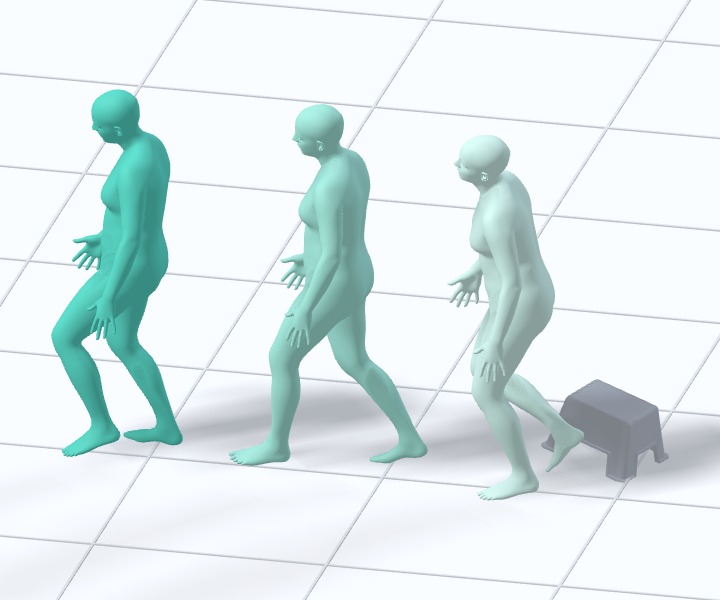} &
\hspace{\scenecmpclip}\includegraphics[width=\scenecmpw]{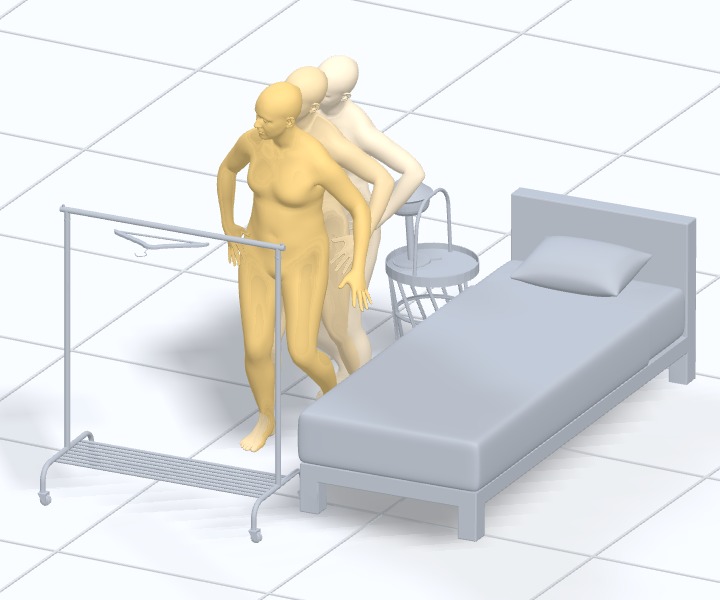} &
\includegraphics[width=\scenecmpw]{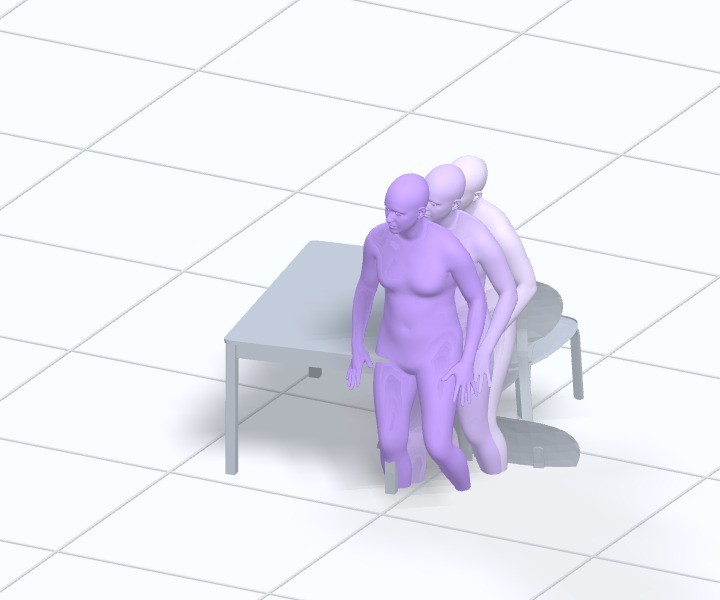} &
\includegraphics[width=\scenecmpw]{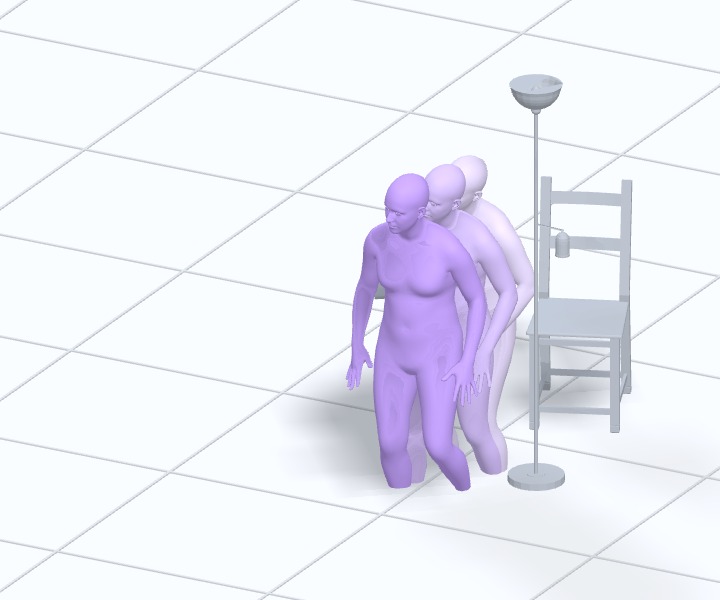} &
\includegraphics[width=\scenecmpw]{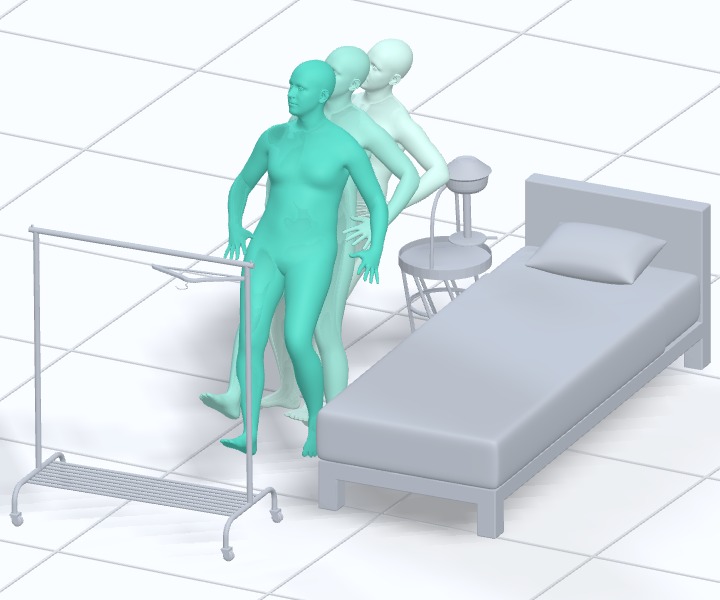}
\end{tabular}\\[1pt]
  \parbox[b][\scenecmph][c]{\scenecmplab}{\centering\rotatebox{90}{\scriptsize\textbf{HiPhi}}}%
\begin{tabular}[b]{@{}cccccccc@{}}
\includegraphics[width=\scenecmpw]{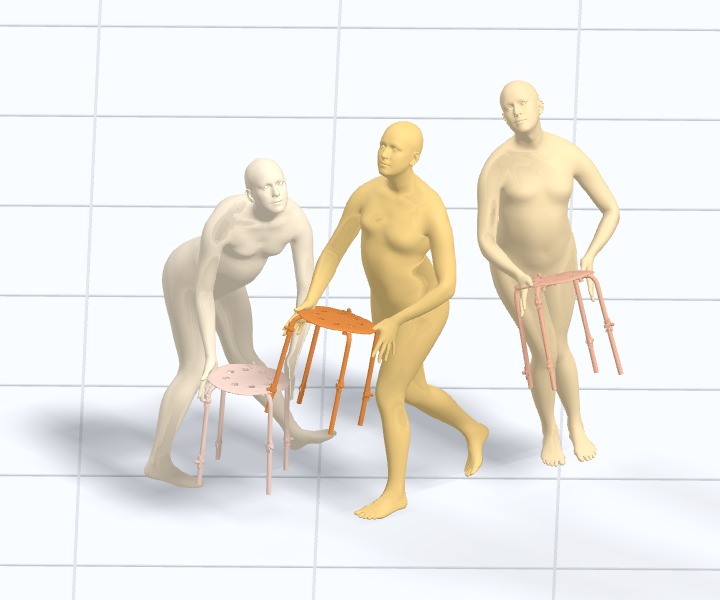} &
\includegraphics[width=\scenecmpw]{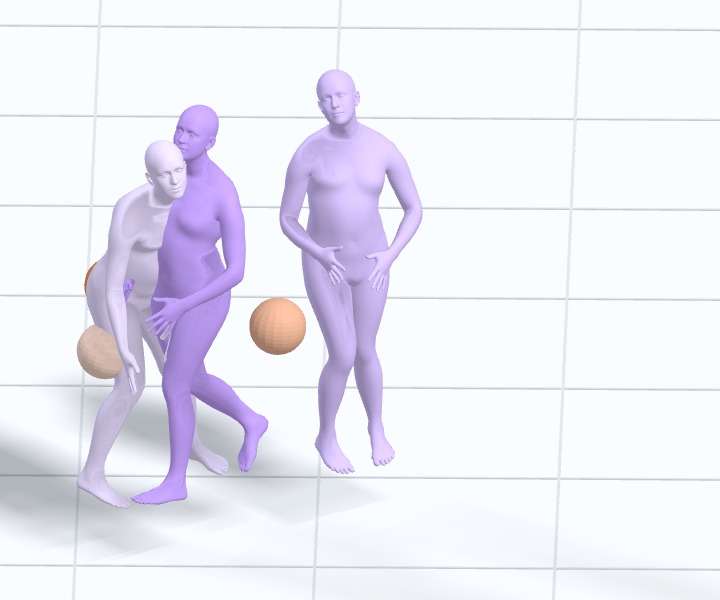} &
\includegraphics[width=\scenecmpw]{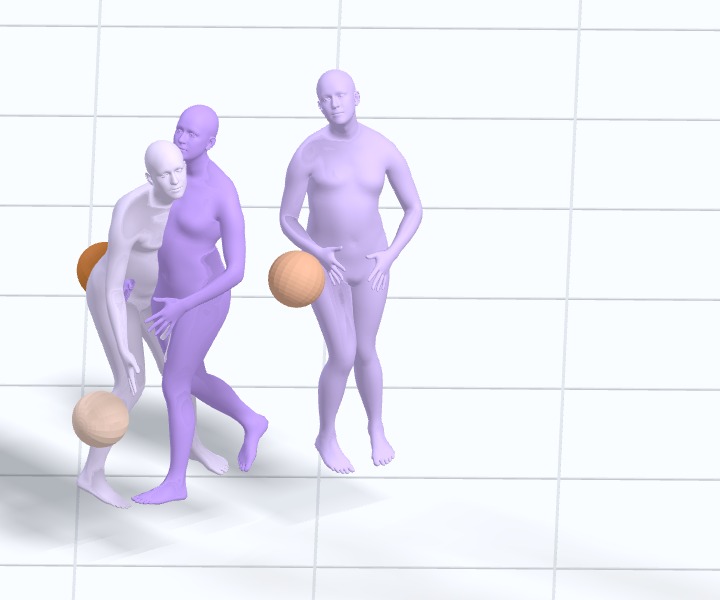} &
\includegraphics[width=\scenecmpw]{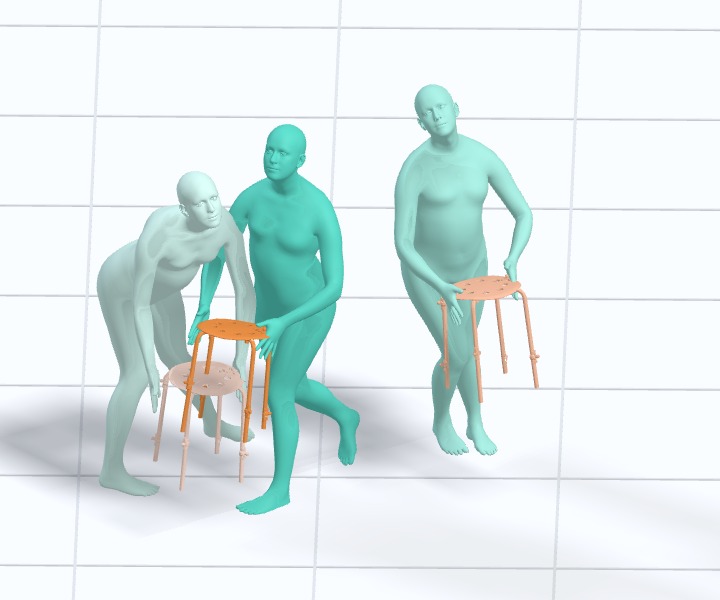} &
\hspace{\scenecmpclip}\includegraphics[width=\scenecmpw]{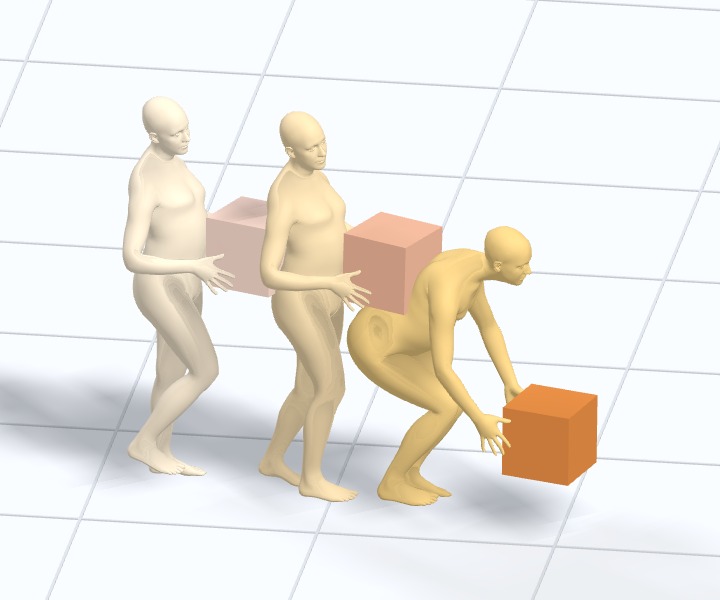} &
\includegraphics[width=\scenecmpw]{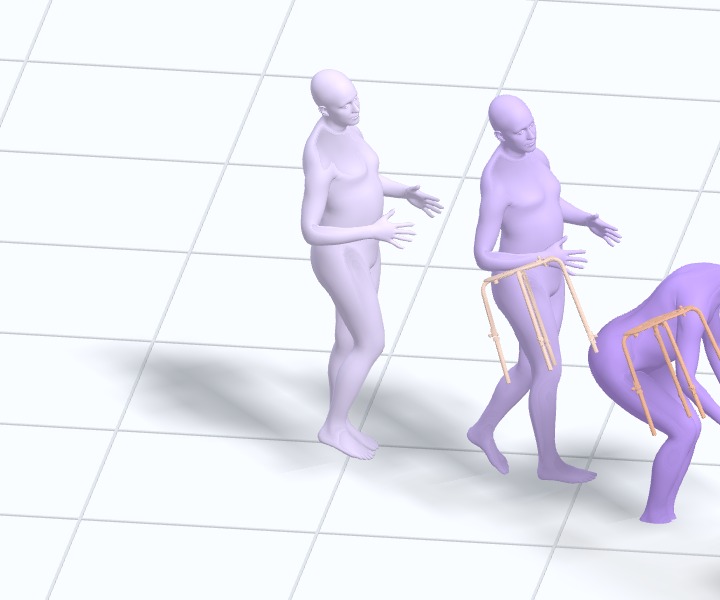} &
\includegraphics[width=\scenecmpw]{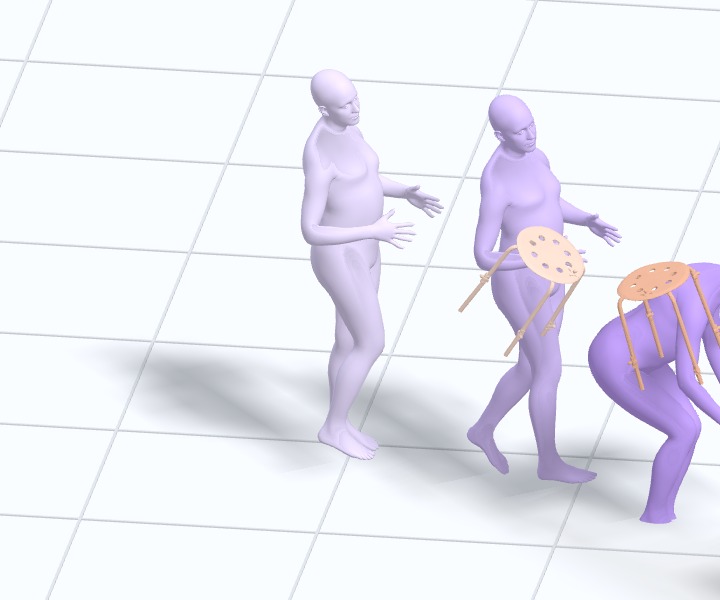} &
\includegraphics[width=\scenecmpw]{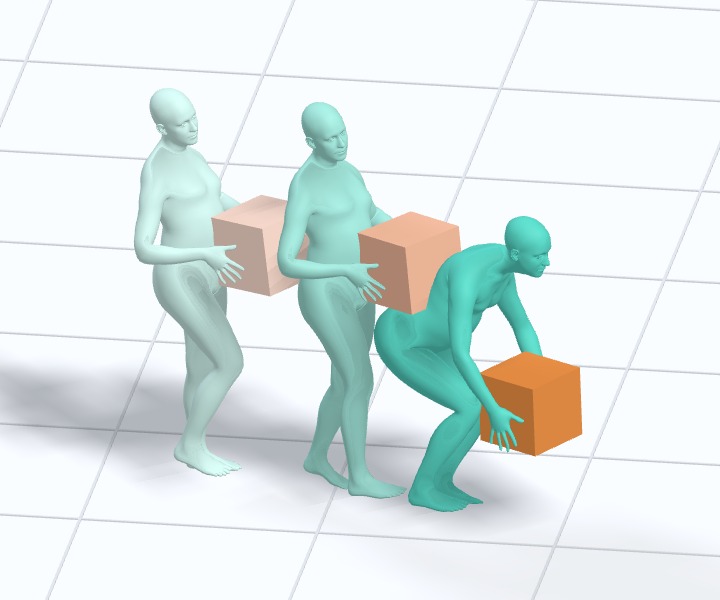}
\end{tabular}\\[1pt]
  \parbox[b][\scenecmph][c]{\scenecmplab}{\centering\rotatebox{90}{\scriptsize\textbf{OMOMO}}}%
\begin{tabular}[b]{@{}cccccccc@{}}
\includegraphics[width=\scenecmpw]{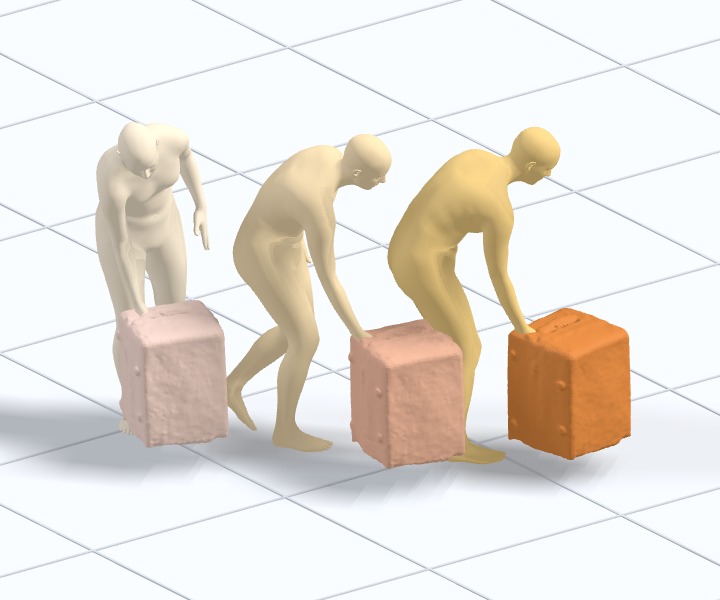} &
\includegraphics[width=\scenecmpw]{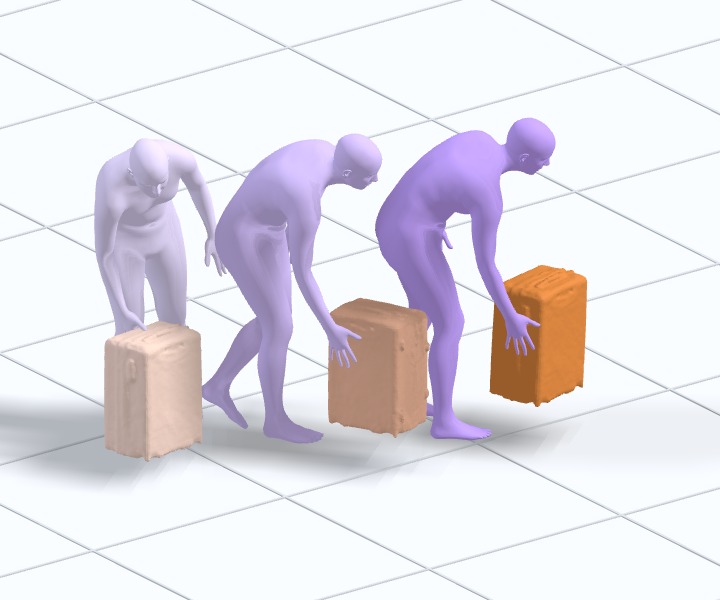} &
\includegraphics[width=\scenecmpw]{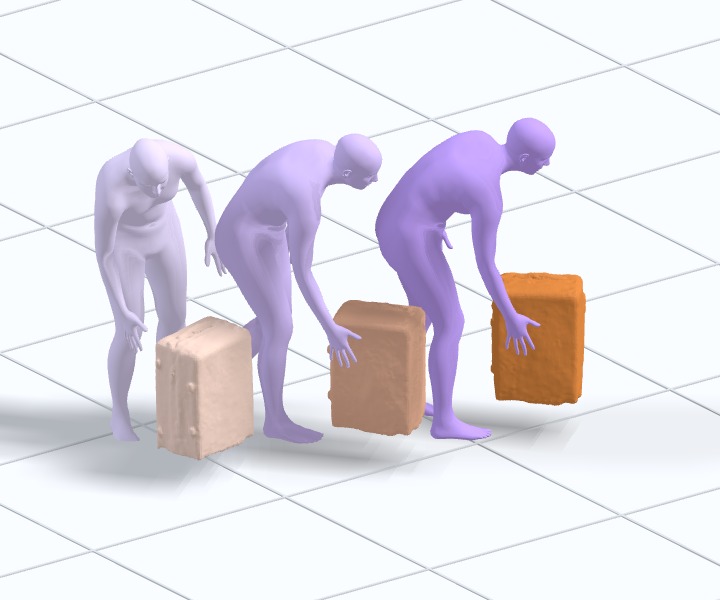} &
\includegraphics[width=\scenecmpw]{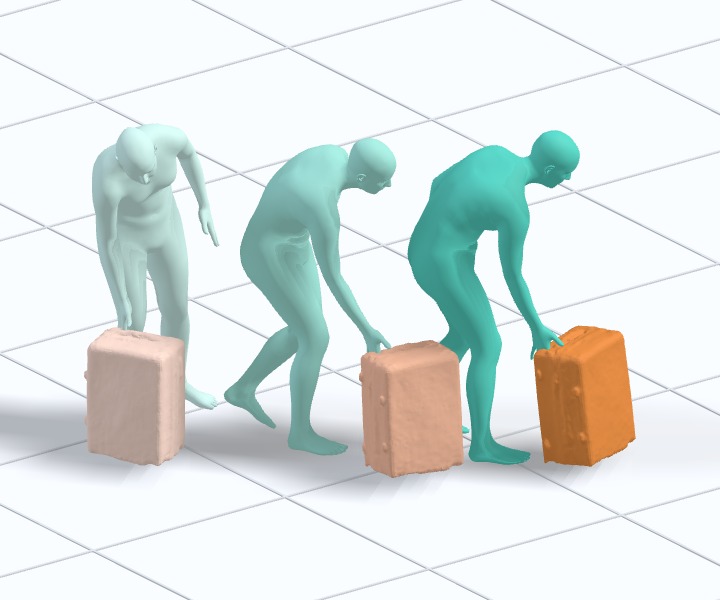} &
\hspace{\scenecmpclip}\includegraphics[width=\scenecmpw]{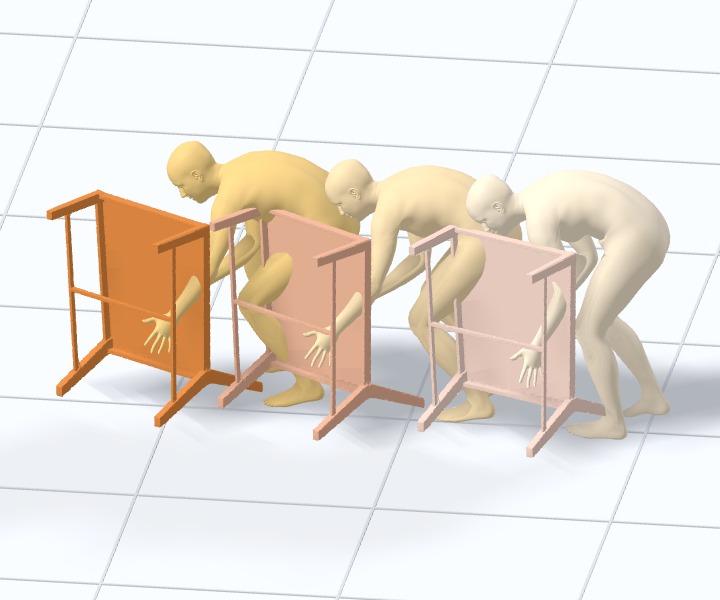} &
\includegraphics[width=\scenecmpw]{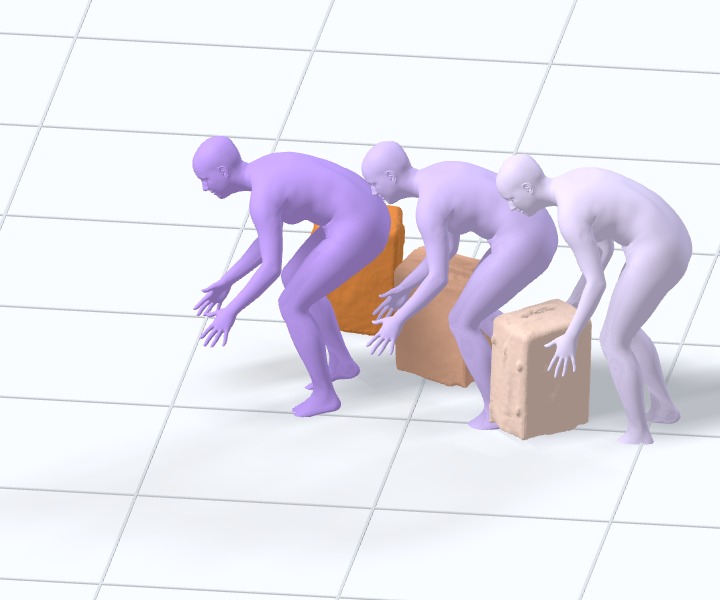} &
\includegraphics[width=\scenecmpw]{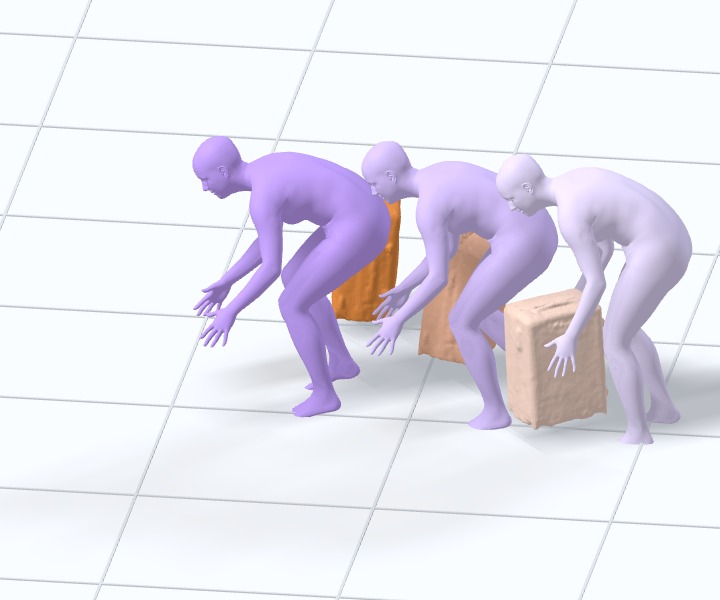} &
\includegraphics[width=\scenecmpw]{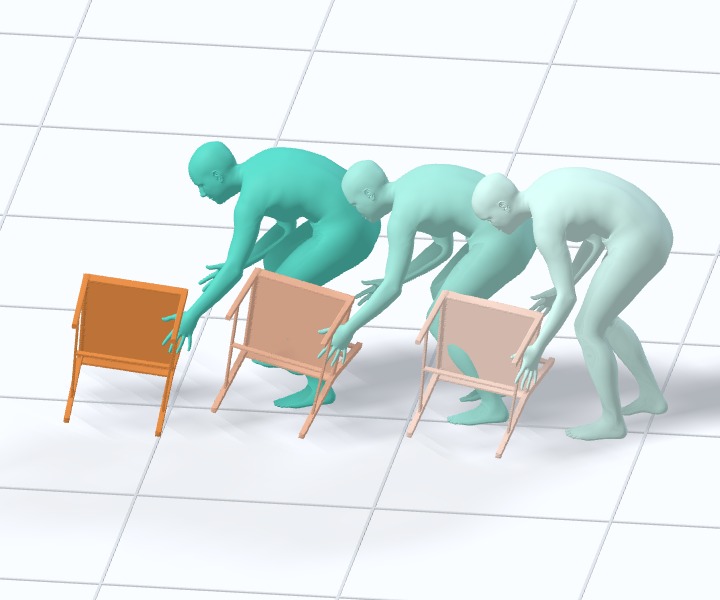}
\end{tabular}
  \vspace{-1mm}
  \caption{\small\textbf{Human--object recovery from IMUs.} Each image overlays three keyframes
  (lighter = earlier) under a shared camera. Bodies: GT (yellow), baselines (purple), Ours (green);
  objects: static (grey), moving (orange). Baselines lift MobilePoser motion to objects: Baseline 1\,/\,2 are
  Pose2Room\,/\,SUMMON on HUMOTO and Object Pop-up\,/\,TriDi on HiPhi and OMOMO.}
  \label{fig:scene_comparison}
\end{figure}

\begin{table*}[t]
\caption{\small\textbf{Dynamic-object results.}
Evaluated with the 5pt setup in the 480-frame setting.
Since Object Pop-up and TriDi take an object category as input, both use the same motion-to-object classifier and therefore share the same recall.}
\centering
\resizebox{0.95\linewidth}{!}{
\begin{tabular}{@{} llccc @{}}
\toprule
Benchmark & Method
& Recall (\%) $\uparrow$
& Moving F1@0.25 (\%) $\uparrow$
& Velocity error (mm/s) $\downarrow$ \\
\midrule
\multirow{3}{*}{OMOMO}
& Mod-I2MTO (Object Pop-up) & 40.42 & 9.73  & 543.87 \\
& Mod-I2MTO (TriDi)     & 40.42 & 13.55 & 1580.35 \\
& Ours                  & \textbf{54.27} & \textbf{24.35} & \textbf{535.37} \\
\midrule
\multirow{3}{*}{HiPhi}
& Mod-I2MTO (Object Pop-up) & 42.70 & 15.33 & 349.59 \\
& Mod-I2MTO (TriDi)     & 42.70 & 16.08 & 1621.91 \\
& Ours                  & \textbf{50.64} & \textbf{26.85} & \textbf{337.97} \\
\bottomrule
\end{tabular}
}
\label{tab:dynamic_object_results}
\end{table*}

\begin{wraptable}{r}{0.55\linewidth}
\centering
\caption{\small\textbf{Quantitative comparison of object predictions in IMU-to-MotionTextObject on HUMOTO~\citep{lu2025humoto}.} Our method consistently outperforms the modular baselines, evaluated on 480-frame IMUs using the 5pt setup (ear, both wrists, both thighs).}
\vspace{-0.5em}
\resizebox{0.85\linewidth}{!}{
\setlength{\tabcolsep}{5pt}
\begin{tabular}{@{} l ccc @{}}
\toprule
Method & 3D-IoU $\uparrow$ & ID-P $\uparrow$ & ID-R $\uparrow$ \\
\midrule
Mod-I2MTO (SUMMON) & 1.65 & 15.22 & 14.14 \\
Mod-I2MTO (Pose2Room) & 4.61 & 8.84 & 16.16 \\
Ours & \textbf{10.75} & \textbf{26.87} & \textbf{36.36} \\
\bottomrule
\end{tabular}
}

\label{tab:quant_imu_to_3d}
\vspace{-0.8em}
\end{wraptable}\noindent{\textbf{IMU-to-MotionText and IMU-to-MotionTextObject.}} We evaluate the I2MT and I2MTO tasks to demonstrate the benefits of jointly reasoning across modalities.
As shown in Tab.~\ref{tab:quant_imu2all}, our model jointly reasons about motion and text from IMU signals, achieving superior performance across all motion and text evaluation metrics compared to a modular baseline Mod-I2MTO that processes IMU signals sequentially through separate single-purpose models (MobilePoser for IMU-to-Motion and MotionGPT3 for motion-to-text). The modular pipeline suffers from error propagation, where inaccuracies in motion prediction cascade into downstream text and object predictions. In contrast, our unified framework learns all modalities within a single model, enabling cross-modal understanding through shared representations.
We further evaluate the quality of object predictions produced by our I2MTO model in Tab.~\ref{tab:quant_imu_to_3d} (static objects), Tab.~\ref{tab:dynamic_object_results} (dynamic objects) and Fig.~\ref{fig:scene_comparison}. To complete the I2MTO baseline, we use either SUMMON~\citep{ye2022scene} or Pose2Room~\citep{nie2022pose2room} on HUMOTO, and Object Pop-up~\citep{petrov2023objpopup} or TriDi~\citep{petrov2025tridi} on OMOMO and HiPhi. Our approach achieves stronger cross-modal consistency compared to baselines, demonstrating the advantage of unified multimodal learning over sequential pipelines.

\noindent{\textbf{Application: Re-localization in a fixed scene.}}  When fine-tuned on sequences captured in a single fixed environment (ParaHome~\citep{kim2025parahome}), our model can additionally predict the person's initial global position and heading from IMU signals alone, which by themselves carry no absolute positional information. 
Please refer to Sec.~\ref{sec:supp_qual_reloc} for more details.

\begin{table*}[t]
\caption{\small\textbf{Ablations on backbone size of our model on HumanML using 5pt inputs and 480 frames.} All variants are trained for 157k iterations, shorter schedule than the full model. Scaling the base model from 355M to 1.3B parameters consistently reduces motion errors (e.g., MPJPE from 36.30 to 23.60), and the largest model also achieves the best text metrics, indicating that our framework benefits from increased model capacity.}
\centering
\resizebox{\linewidth}{!}{
\setlength{\tabcolsep}{2pt}
\begin{tabular}{@{} l c ccccc ccccc @{}}
\toprule
\multirow{2}{*}{Base model} & \multirow{2}{*}{\#Params} & \multicolumn{5}{c}{Motion Metrics} & \multicolumn{5}{c}{Text Metrics} \\
\cmidrule(lr){3-7} \cmidrule(lr){8-12}
& & MPJPE $\downarrow$ & PA-MPJPE $\downarrow$ & MPJRE $\downarrow$ & MPJVE $\downarrow$ & MTE $\downarrow$
& ROUGE-L $\uparrow$ & CIDEr $\uparrow$ & BLEU@1 $\uparrow$ & BLEU@4 $\uparrow$ & BERT $\uparrow$ \\
\midrule
GPT-2 medium & 355M & 36.30 & 27.15 & 7.13 & 47.16 & 115.20 & 41.11 & 169.25 & 35.32 & 7.68 & 31.76 \\
Qwen3-0.6B & 0.6B & 26.52 & 19.40 & 5.52 & 34.62 & 67.49 & 39.88 & 165.54 & 34.01 & 7.49 & 30.34 \\
Phi-1.5 & 1.3B & \textbf{23.60} & \textbf{17.66} & \textbf{4.99} & \textbf{30.59} & \textbf{63.55} & \textbf{42.15} & \textbf{180.09} & \textbf{36.63} & \textbf{8.78} & \textbf{33.71} \\
\bottomrule
\end{tabular}
}
\label{tab:scaling_laws}
\end{table*}

\begin{table*}[t]
\caption{\small\textbf{Joint vs.\ single-task and cascaded formulations.}  All variants share the same architecture and training setup, using 480-frame IMUs under the 5pt setup (ear, both wrists, both thighs). Single-task variants predict motion (I2M), text (I2T), or objects (I2O) directly from IMUs; cascaded variants (I2M + M2T / M2O) feed the motion predicted by I2M into a motion-to-text (M2T) or motion-to-object (M2O) model. Our joint formulation (I2MTO) consistently outperforms the cascaded (I2M + M2T / M2O) and single-task (I2M / I2T / I2O) approaches across all text and object metrics and most motion metrics, demonstrating that joint reasoning not only mitigates error propagation in cascaded pipelines but also improves individual task performance.}
\centering
\small
\setlength{\tabcolsep}{2pt}
\renewcommand{\arraystretch}{0.90}
\captionsetup[subtable]{skip=1pt}

\begin{subtable}[t]{0.6\linewidth}
\centering
\caption{IMU-to-Motion on HumanML}
\resizebox{\linewidth}{!}{
\begin{tabular}{@{} l ccccc @{}}
\toprule
Method & MPJPE $\downarrow$ & PA-MPJPE $\downarrow$ & MPJRE $\downarrow$ & MPJVE $\downarrow$ & MTE $\downarrow$ \\
\midrule
I2M & 16.76 & 12.59 & 3.65 & 21.69 & \textbf{44.28} \\
I2MTO (Ours) & \textbf{14.69} & \textbf{11.32} & \textbf{3.06} & \textbf{19.31} & 48.50 \\
\bottomrule
\end{tabular}}
\end{subtable}

\vspace{2pt}

\begin{subtable}[t]{0.62\linewidth}
\centering
\caption{IMU-to-Text on HumanML}
\resizebox{\linewidth}{!}{
\begin{tabular}{@{} l ccccc @{}}
\toprule
Method & ROUGE-L $\uparrow$ & CIDEr $\uparrow$ & BLEU@1 $\uparrow$ & BLEU@4 $\uparrow$ & BERT $\uparrow$ \\
\midrule
I2T & 37.28 & 137.20 & 30.67 & 3.98 & 25.79 \\
I2M + M2T & 35.28 & 127.70 & 28.95 & 3.37 & 21.23 \\
I2MTO (Ours) & \textbf{43.35} & \textbf{197.96} & \textbf{38.08} & \textbf{11.00} & \textbf{35.17} \\
\bottomrule
\end{tabular}}
\end{subtable}
\hfill
\begin{subtable}[t]{0.36\linewidth}
\centering
\caption{IMU-to-Object on HUMOTO}
\resizebox{\linewidth}{!}{
\begin{tabular}{@{} l ccc @{}}
\toprule
Method & 3D-IoU $\uparrow$ & ID-P $\uparrow$ & ID-R $\uparrow$ \\
\midrule
I2O & 1.76 & 14.71 & 15.15 \\
I2M + M2O & 1.09 & 20.93 & 18.18 \\
I2MTO (Ours) & \textbf{10.75} & \textbf{26.87} & \textbf{36.36} \\
\bottomrule
\end{tabular}}
\end{subtable}

\label{tab:joint_formulations}
\end{table*}

\subsection{Analysis}
We summarize the answers to the three questions posed at the beginning of this section.

\noindent\textbf{(1) Accuracy of \ourabbr.} \ourabbr improves motion and captioning over all baselines on both synthetic and real-world IMU data (Tabs.~\ref{tab:quant_imu2motion}--\ref{tab:real_world_commodity}). For objects, it recovers substantially more of the interacted objects and better captures the motion of dynamic objects (Tabs.~\ref{tab:dynamic_object_results} and~\ref{tab:quant_imu_to_3d}), although absolute 3D localization remains challenging.

\noindent\textbf{(2) Benefits of large pre-training.} As shown in Tab.~\ref{tab:scaling_laws}, both I2M and I2MT scale effectively with model capacity. On the interaction datasets, compared with direct training from Show-o, large-scale motion--text pretraining helps motion the most (Tab.~\ref{tab:stage1_pretraining}). Text metrics are similar, as captions in these datasets have low diversity and are easy to fit; static objects are comparable, while dynamic objects are better captured with pretraining (see Appendix~\ref{sec:supp_stage1_pretraining} for details).

\begin{wraptable}{r}{0.36\linewidth}
\centering
\vspace{-1.0em}
\caption{\small\textbf{Effect of large-scale motion--text pretraining (PT)} compared with direct training from Show-o for the same iterations (5pt, 480 frames).}
\vspace{-0.6em}
\resizebox{\linewidth}{!}{
\setlength{\tabcolsep}{3pt}
\begin{tabular}{@{} l c ccc @{}}
\toprule
Benchmark & PT & MPJPE $\downarrow$ & MTE $\downarrow$ & BERT $\uparrow$ \\
\midrule
\multirow{2}{*}{HUMOTO} & $\times$ & 63.80 & 230.54 & 20.50 \\
 & $\checkmark$ & \textbf{31.41} & \textbf{61.67} & \textbf{22.87} \\
\midrule
\multirow{2}{*}{OMOMO} & $\times$ & 43.45 & 116.39 & 63.17 \\
 & $\checkmark$ & \textbf{31.25} & \textbf{59.23} & \textbf{63.45} \\
\midrule
\multirow{2}{*}{HiPhi} & $\times$ & 30.26 & 80.68 & 23.63 \\
 & $\checkmark$ & \textbf{21.59} & \textbf{52.11} & \textbf{24.66} \\
\bottomrule
\end{tabular}
}
\label{tab:stage1_pretraining}
\vspace{-1.2em}
\end{wraptable}
\noindent\textbf{(3) Joint reasoning vs.\ cascaded approaches.} Our model outperforms the cascade of state-of-the-art modules on all motion and text metrics (Tab.~\ref{tab:quant_imu2all}) and in object recall (Tabs.~\ref{tab:dynamic_object_results} and~\ref{tab:quant_imu_to_3d}). To rule out differences in architecture and pretraining, Tab.~\ref{tab:joint_formulations} compares single-task, cascaded, and joint training with the same model: joint training is best on all text and object metrics and on four of the five motion metrics, with the largest gain on object prediction, where interaction data are scarce.

\noindent\textbf{(4) Challenges and limitations.}
Attention alone cannot fully prevent human--object penetration; explicit contact losses are left for future work. We recover interaction-relevant objects rather than the full 3D scene, whose reconstruction from sparse IMUs would require substantial hallucination. Object prediction is limited to a closed set of categories seen in training and does not recover object shape; absolute localization accuracy (e.g., 3D-IoU) remains low, reflecting the inherent ambiguity of inferring object placement from body motion alone.

\section{Conclusion}

We introduced \ourabbr, a framework for 4D human--object understanding without relying on cameras. By repurposing large language models for non-visual spatiotemporal reasoning, our approach recovers human motion, activity descriptions, and interacted objects using only a few everyday wearable IMUs. Experiments across diverse real-world datasets demonstrate that \ourabbr produces more coherent and temporally stable estimates than state-of-the-art cascaded pipelines. These results highlight the strong potential of wearable sensors as a privacy-preserving and scalable alternative to vision-based perception.

\subsubsection*{Acknowledgments}
This project is supported by the Amazon Illinois AICE Center Research Grant. Hao-Yu Hsu is supported by the Amazon AI PhD Scholarship. This work used the DeltaAI system~\citep{bode2025deltaai} at the National Center for Supercomputing Applications (NSF award OAC 2320345 and the State of Illinois) through allocation CIS251344 from the ACCESS program, which is supported by NSF grants \#2138259, \#2138286, \#2138307, \#2137603, and \#2138296. We also thank Prof. Romit Roy Choudhury, Dr. Robinson Piramuthu, and Dr. Gunnar Sigurdsson for their helpful discussions.

\clearpage
\bibliography{main}
\bibliographystyle{iclr2027_conference}

\clearpage
\appendix

In this appendix, we provide additional results and implementation details. It is organized as follows:
\begin{itemize}
\item Sec.~\ref{sec:supp_baseline} provides additional baseline comparisons for the IMU-to-Motion task, as well as alternative modular pipeline configurations used as baselines for the IMU-to-MotionText task.
\item Sec.~\ref{sec:supp_ablation_studies} includes ablation studies on input configurations and model designs.
\item Sec.~\ref{sec:supp_qualitative} demonstrates diverse outputs through qualitative results.
\item Sec.~\ref{sec:supp_implementation_details} provides implementation details for both the training and testing phases.
\item Sec.~\ref{sec:supp_dataset_details} describes the datasets used for Stage-1 pretraining and dataset-specific finetuning.
\item Sec.~\ref{sec:supp_evaluation_details} details the evaluations.
\item Sec.~\ref{sec:supp_failure} illustrates failure cases and limitations.
\end{itemize}

\section{More Baseline Comparisons}
\label{sec:supp_baseline}

In this section, we provide comparisons to additional baselines of IMU-to-Motion and IMU-to-MotionText.

\subsection{IMU-to-Motion}
In Tab.~\ref{tab:quant_imu2motion} and Tab.~\ref{tab:real_world_commodity} of the main paper, we evaluate our method under the 3-sensor setting (3pt) on real-world IMU data from DIP-IMU, IMUPoser, and our IMU4D-Real dataset. We additionally compare to ProgIP~\citep{zhu2025progressive}, IMUPoser and MobilePoser on LINGO dataset in both 3pt and 5pt settings in Tab.~\ref{tab:quant_imu2motion_lingo_3pt}.
\begin{table*}[h]
\caption{\small\textbf{Quantitative comparison of IMU-to-Motion on LINGO~\citep{jiang2024autonomous}.} Evaluated on 480 frames. *ProgIP uses a 3-point head-and-wrists configuration. N/A: the method does not predict global translation.}
\centering
\resizebox{0.9\linewidth}{!}{
\setlength{\tabcolsep}{5pt}
\begin{tabular}{@{} l l ccccc @{}}
\toprule
Method & \#IMUs & MPJPE $\downarrow$ & PA-MPJPE $\downarrow$ & MPJRE $\downarrow$ & MPJVE $\downarrow$ & MTE $\downarrow$ \\
\midrule
ProgIP & 3pt* & 64.89 & 45.66 & 15.52 & 68.77 & N/A \\
IMUPoser & 3pt & 55.46 & 43.86 & 14.70 & 74.17 & N/A \\
MobilePoser & 3pt & 59.75 & 49.67 & 18.53 & 78.61 & 208.06 \\
Ours & 3pt & \textbf{40.67} & \textbf{28.54} & \textbf{10.21} & \textbf{52.28} & \textbf{28.87} \\
\midrule
IMUPoser & 5pt & 46.69 & 35.16 & 13.28 & 62.49 & N/A \\
MobilePoser & 5pt & 49.61 & 39.63 & 17.23 & 64.95 & 195.95 \\
Ours & 5pt & \textbf{31.53} & \textbf{18.55} & \textbf{8.13} & \textbf{40.56} & \textbf{24.39} \\
\bottomrule
\end{tabular}
}
\label{tab:quant_imu2motion_lingo_3pt}
\end{table*}

\subsection{Alternative Modular Pipeline Configurations in IMU-to-MotionText}
In the main paper, we use a cascaded pipeline of MobilePoser (IMU$\to$motion) and MotionGPT3 (motion$\to$text) as the baseline for IMU-to-MotionText. To show the effectiveness of joint modeling over cascaded approaches, we further compare additional combinations of state-of-the-art IMU-to-motion methods (IMUPoser~\citep{mollyn2023imuposer}, MobilePoser~\citep{xu2024mobileposer}) and motion-to-text methods (MotionGPT~\citep{jiang2023motiongpt}, MotionGPT3~\citep{zhu2025motiongpt3}, GeoMotionGPT~\citep{ye2026geomotiongpt}) as modular pipelines. 

As shown in Table~\ref{tab:quant_imu_to_text}, our unified model consistently outperforms all modular combinations across all text metrics on the HumanML dataset, validating the benefits of joint multimodal generation over cascaded approaches. Although IMUPoser with MotionGPT3 yields slightly better captions than MobilePoser with MotionGPT3, we build the main-paper baseline on MobilePoser because it also predicts the global translation required by the downstream object predictors, whereas IMUPoser predicts only local joint rotations.

\begin{table}[t]
\caption{\textbf{Ablation on alternative IMU-to-motion and motion-to-text methods on HumanML.} Evaluated using 5pt inputs and 480 frames. All IMU-to-motion and motion-to-text baselines are trained on the same datasets as ours, and our unified model outperforms every combination of them across all text metrics.}
\centering
\resizebox{\linewidth}{!}
{
\setlength{\tabcolsep}{5pt}
\begin{tabular}{@{} l l | ccccc @{}}
\toprule
IMU$\rightarrow$Motion & Motion$\rightarrow$Text & ROUGE-L $\uparrow$ & CIDEr $\uparrow$ & BLEU@1 $\uparrow$ & BLEU@4 $\uparrow$ & BERT $\uparrow$ \\
\midrule

IMUPoser & MotionGPT & 28.94 & 86.40 & 24.72 & 2.09 & 13.89 \\
IMUPoser & MotionGPT3 & 37.84 & 141.71 & 33.08 & 5.13 & 27.87 \\
IMUPoser & GeoMotionGPT & 27.11 & 76.16 & 22.51 & 1.33 & 11.88 \\
MobilePoser & MotionGPT & 29.69 & 85.26 & 25.21 & 2.39 & 15.56 \\
MobilePoser & MotionGPT3 & 36.80 & 133.15 & 32.55 & 4.85 & 26.50 \\
MobilePoser & GeoMotionGPT & 30.68 & 98.89 & 25.99 & 2.90 & 17.05 \\
\multicolumn{2}{c|}{Ours} & \textbf{43.35} & \textbf{197.96} & \textbf{38.08} & \textbf{11.00} & \textbf{35.17} \\
\bottomrule
\end{tabular}
}
\label{tab:quant_imu_to_text}
\end{table}
\subsection{Baseline Implementation Details}
\label{sec:supp_baseline_impl}
We build all cascaded baselines on their official implementations and adapt them to our data with minimal changes. Each baseline is trained separately on each dataset with the same training split as ours, and checkpoints are selected on a held-out portion of the training set. Since the original methods use SMPL or other body models, we represent all bodies with a neutral SMPL-X model~\citep{pavlakos2019expressive} and express object poses in the same canonical object frames as our model. At test time, every baseline operates on the motion predicted by MobilePoser.

\noindent{\textbf{Motion-to-text.}} We retrain MotionGPT3~\citep{zhu2025motiongpt3} on our motion--text training data, converting each motion into the 272-dimensional representation of~\citet{xiao2025motionstreamer} that the model takes as input.

\noindent{\textbf{Static object baselines.}} SUMMON~\citep{ye2022scene} and Pose2Room~\citep{nie2022pose2room} predict a static object layout from motion. For SUMMON, we retrain its contact predictor with the object categories of each dataset and reimplement its object placement step, which fits candidate object meshes to the predicted contacts while penalizing body penetration; we disable its scene-completion stage, which targets full rooms. As the original conditional VAE predicted no contacts at test time on our data, we train it with a deterministic latent. For Pose2Room, we adapt its skeleton graph to the SMPL-X body joints and adjust its voting radii and rotation parameterization, which were designed for furniture, to also handle small handheld objects.

\noindent{\textbf{Dynamic object baselines.}} Object Pop-up~\citep{petrov2023objpopup} and TriDi~\citep{petrov2025tridi} predict an object's pose from the human body at individual frames, which we interpolate to 30\,fps to obtain object trajectories. Both methods originally condition on the ground-truth object identity. To avoid this oracle, we train a motion-to-object classifier, a bidirectional LSTM on body joint trajectories, and condition both methods on its prediction; as a result, they share the same recall in Tab.~\ref{tab:dynamic_object_results}. Since the training code of TriDi's contact autoencoder is not publicly available, we reimplement it. Neither method detects objects, so the number of objects in each sequence is taken from the ground truth, which favors these baselines.

\section{Ablation Studies}
\label{sec:supp_ablation_studies}

\subsection{LLM Initialization}

We ablate the importance of initializing our model with pretrained LLM weights by comparing against training the same architecture from scratch in Tab.~\ref{tab:llm_initialization}. LLM initialization consistently improves performance across all motion and text metrics. Notably, the benefits extend beyond text generation to motion reconstruction, reducing MPJPE from 25.77 to 23.60 and MTE from 71.89 to 63.55. These results suggest that the representations learned during pretraining provide a strong initialization not only for text understanding and generation, but also for learning the mapping between IMU signals and human motion.

\begin{table*}[t]
\caption{\small\textbf{Ablation on initialization from an LLM.} Evaluated on HumanML using 5pt inputs and 480 frames, with both variants trained for 157k iterations.}
\centering
\resizebox{\linewidth}{!}{
\setlength{\tabcolsep}{2pt}
\begin{tabular}{@{} c ccccc ccccc @{}}
\toprule
\multirow{2}{*}{Init. from LLM?} & \multicolumn{5}{c}{Motion Metrics} & \multicolumn{5}{c}{Text Metrics} \\
\cmidrule(lr){2-6} \cmidrule(lr){7-11}
& MPJPE $\downarrow$ & PA-MPJPE $\downarrow$ & MPJRE $\downarrow$ & MPJVE $\downarrow$ & MTE $\downarrow$
& ROUGE-L $\uparrow$ & CIDEr $\uparrow$ & BLEU@1 $\uparrow$ & BLEU@4 $\uparrow$ & BERT $\uparrow$ \\
\midrule
$\times$ & 25.77 & 19.32 & 5.38 & 33.67 & 71.89 & 41.70 & 175.56 & 36.12 & 8.09 & 32.69 \\
$\checkmark$ & \textbf{23.60} & \textbf{17.66} & \textbf{4.99} & \textbf{30.59} & \textbf{63.55} & \textbf{42.15} & \textbf{180.09} & \textbf{36.63} & \textbf{8.78} & \textbf{33.71} \\
\bottomrule
\end{tabular}
}
\label{tab:llm_initialization}
\end{table*}

\subsection{Large-Scale Motion--Text Pretraining for Interaction Datasets}
\label{sec:supp_stage1_pretraining}

To isolate the effect of large-scale motion--text pretraining, we compare our model, which is finetuned on each human--object interaction dataset from the pretrained checkpoint, with direct training from Show-o on the same dataset, \ie, the same LLM initialization without large-scale pretraining. Both variants use the same architecture, objective, and number of training steps (50k on HUMOTO and OMOMO, 100k on HiPhi; Sec.~\ref{sec:supp_implementation_details}). Tab.~\ref{tab:stage1_pretraining_full} reports the full results.

\noindent{\textbf{Motion.}} Pretraining helps motion the most. It consistently reduces all motion errors on every dataset, with the largest gain on HUMOTO, the smallest of the three datasets, where in-domain data alone are insufficient to learn an accurate motion prior.

\noindent{\textbf{Text.}} Text metrics are similar with and without pretraining. Since captions in these interaction datasets have low diversity, they are easy to fit from in-domain data alone, leaving little room for pretraining to help.

\noindent{\textbf{Objects.}} For static objects on HUMOTO, localization and identification are close between the two variants. For dynamic objects, pretraining captures object motion better, as object trajectories are decoded from the motion queries and benefit from more accurate human motion.

\begin{table*}[t]
\caption{\small\textbf{Effect of large-scale motion--text pretraining on human--object interaction datasets.} PT indicates whether the model is finetuned from the pretrained checkpoint ($\checkmark$, Ours) or trained directly from Show-o ($\times$). Both variants are trained for 50k steps on HUMOTO and OMOMO and 100k steps on HiPhi, and evaluated on 480-frame IMUs using the 5pt setup (ear, both wrists, both thighs). Static-object metrics apply to HUMOTO and dynamic-object metrics to OMOMO and HiPhi; --: not applicable.}
\centering
\captionsetup[subtable]{skip=2pt}

\begin{subtable}[t]{\linewidth}
\centering
\caption{Motion and text}
\resizebox{\linewidth}{!}{
\setlength{\tabcolsep}{3pt}
\begin{tabular}{@{} l c ccccc ccccc @{}}
\toprule
\multirow{2}{*}{Benchmark} & \multirow{2}{*}{PT} & \multicolumn{5}{c}{Motion Metrics} & \multicolumn{5}{c}{Text Metrics} \\
\cmidrule(lr){3-7} \cmidrule(lr){8-12}
& & MPJPE $\downarrow$ & PA-MPJPE $\downarrow$ & MPJRE $\downarrow$ & MPJVE $\downarrow$ & MTE $\downarrow$ & ROUGE-L $\uparrow$ & CIDEr $\uparrow$ & BLEU@1 $\uparrow$ & BLEU@4 $\uparrow$ & BERT $\uparrow$ \\
\midrule
\multirow{2}{*}{HUMOTO} & $\times$ & 63.80 & 57.38 & 12.16 & 84.21 & 230.54 & 33.83 & 101.56 & 34.43 & \textbf{7.23} & 20.50 \\
 & $\checkmark$ & \textbf{31.41} & \textbf{25.60} & \textbf{7.51} & \textbf{38.16} & \textbf{61.67} & \textbf{34.09} & \textbf{104.85} & \textbf{35.52} & 5.79 & \textbf{22.87} \\
\midrule
\multirow{2}{*}{OMOMO} & $\times$ & 43.45 & 33.84 & 8.78 & 60.27 & 116.39 & \textbf{80.51} & \textbf{633.84} & \textbf{79.02} & \textbf{61.68} & 63.17 \\
 & $\checkmark$ & \textbf{31.25} & \textbf{22.16} & \textbf{6.93} & \textbf{44.53} & \textbf{59.23} & 79.74 & 619.57 & 78.14 & 59.63 & \textbf{63.45} \\
\midrule
\multirow{2}{*}{HiPhi} & $\times$ & 30.26 & 23.49 & 8.11 & 38.72 & 80.68 & 40.80 & 166.22 & 41.07 & 11.29 & 23.63 \\
 & $\checkmark$ & \textbf{21.59} & \textbf{15.70} & \textbf{6.96} & \textbf{29.29} & \textbf{52.11} & \textbf{41.66} & \textbf{186.81} & \textbf{41.21} & \textbf{14.16} & \textbf{24.66} \\
\bottomrule
\end{tabular}}
\end{subtable}

\vspace{6pt}

\begin{subtable}[t]{\linewidth}
\centering
\caption{Objects}
\resizebox{0.68\linewidth}{!}{
\setlength{\tabcolsep}{3pt}
\begin{tabular}{@{} l c ccc ccc @{}}
\toprule
\multirow{2}{*}{Benchmark} & \multirow{2}{*}{PT} & \multicolumn{3}{c}{Static Objects} & \multicolumn{3}{c}{Dynamic Objects} \\
\cmidrule(lr){3-5} \cmidrule(lr){6-8}
& & 3D-IoU $\uparrow$ & ID-P $\uparrow$ & ID-R $\uparrow$ & Recall $\uparrow$ & Moving F1 $\uparrow$ & Vel.\ err.\ $\downarrow$ \\
\midrule
\multirow{2}{*}{HUMOTO} & $\times$ & \textbf{10.84} & 22.63 & 31.31 & -- & -- & -- \\
 & $\checkmark$ & 10.75 & \textbf{26.87} & \textbf{36.36} & -- & -- & -- \\
\midrule
\multirow{2}{*}{OMOMO} & $\times$ & -- & -- & -- & 48.50 & 14.65 & 615.15 \\
 & $\checkmark$ & -- & -- & -- & \textbf{54.27} & \textbf{24.35} & \textbf{535.37} \\
\midrule
\multirow{2}{*}{HiPhi} & $\times$ & -- & -- & -- & \textbf{56.01} & 23.34 & 383.60 \\
 & $\checkmark$ & -- & -- & -- & 50.64 & \textbf{26.85} & \textbf{337.97} \\
\bottomrule
\end{tabular}}
\end{subtable}

\label{tab:stage1_pretraining_full}
\end{table*}

\subsection{Sensor Count and Placement}
\noindent{\textbf{Sensor placement.}} We evaluate the robustness of our model to different sensor placements on HumanML in Tab.~\ref{tab:sensor_placement_humanml}. Specifically, we consider all eight left/right combinations of the wrist, thigh, and ear sensors under the 3pt setup. Performance remains highly consistent across these configurations for both motion reconstruction and text generation. In particular, MPJPE varies by only $\pm0.16$\,mm across placements, while similarly small variations are observed for the text metrics. These results demonstrate that our model is robust to which side of the body each sensor is worn, providing flexibility in sensor placement for practical deployment. The same holds with fewer sensors: moving the thigh sensor to the other side under the 2pt setup changes MPJPE by only 0.25\,mm (Tab.~\ref{tab:sensor_count_humanml}).

\noindent{\textbf{Number of sensors.}} Tab.~\ref{tab:sensor_count_humanml} shows how performance changes as sensors are removed one at a time, starting from the full 5pt setup. Motion error increases steadily (MPJPE 23.6 / 29.4 / 38.0 / $\sim$60\,mm for 5/3/2/1pt), as expected from the loss of body constraints. Text generation is much less sensitive: 3pt keeps 99\% of the 5pt CIDEr, and even a phone-and-watch 2pt setup keeps 91\% (164.2 vs.\ 180.1), showing that a coarse sense of the activity survives when the motion itself is only partially observed. With a single sensor both tasks degrade clearly. Which sensor is kept also matters: a wrist-only input has no reading near the pelvis, so the global trajectory drifts (MTE 289\,mm vs.\ 93\,mm for thigh-only), and its captions are also worse (CIDEr 125 vs.\ 139). This suggests that a pocket-worn phone is the more informative single device.

\begin{table*}[t]
\caption{\small\textbf{Sensor-placement variations on HumanML.} Evaluated on 480 frames using the 3pt setup (one wrist, one thigh, and one ear); (L) and (R) denote the left and right side of the body. The last two rows report the mean and sample standard deviation across all eight placements. Performance is remarkably stable: the standard deviation across placements is only 0.16\,mm for MPJPE and below 0.25 points for every text metric except CIDEr ($\pm$1.33), indicating that our model does not depend on which side of the body the sensors are worn.}
\centering
\resizebox{\linewidth}{!}{
\setlength{\tabcolsep}{2pt}
\begin{tabular}{@{} l ccccc ccccc @{}}
\toprule
\multirow{2}{*}{IMU placement} & \multicolumn{5}{c}{Motion Metrics} & \multicolumn{5}{c}{Text Metrics} \\
\cmidrule(lr){2-6} \cmidrule(lr){7-11}
& MPJPE $\downarrow$ & PA-MPJPE $\downarrow$ & MPJRE $\downarrow$ & MPJVE $\downarrow$ & MTE $\downarrow$
& ROUGE-L $\uparrow$ & CIDEr $\uparrow$ & BLEU@1 $\uparrow$ & BLEU@4 $\uparrow$ & BERT $\uparrow$ \\
\midrule
(L) wrist, (L) thigh, (L) ear & 29.35 & 24.16 & 5.82 & 38.78 & \textbf{70.56} & \textbf{41.69} & 178.27 & 36.15 & 8.65 & 32.96 \\
(L) wrist, (R) thigh, (L) ear & 29.44 & 24.24 & 5.82 & 38.72 & 71.46 & 41.51 & 175.89 & 35.92 & 8.55 & 32.99 \\
(R) wrist, (L) thigh, (L) ear & \textbf{29.12} & \textbf{23.76} & \textbf{5.80} & \textbf{38.38} & 70.94 & 41.39 & 175.68 & 35.91 & 8.40 & 32.62 \\
(R) wrist, (R) thigh, (L) ear & 29.21 & 23.90 & 5.82 & 38.43 & 71.81 & 41.50 & 177.66 & 36.04 & 8.68 & 32.94 \\
(L) wrist, (L) thigh, (R) ear & 29.52 & 24.29 & 5.85 & 39.09 & 71.30 & 41.62 & 177.09 & 35.99 & 8.56 & 32.87 \\
(L) wrist, (R) thigh, (R) ear & 29.61 & 24.36 & 5.86 & 39.02 & 72.11 & 41.64 & \textbf{178.50} & \textbf{36.23} & \textbf{8.70} & \textbf{33.05} \\
(R) wrist, (L) thigh, (R) ear & 29.26 & 23.86 & 5.83 & 38.62 & 71.10 & 41.58 & 177.11 & 36.02 & 8.58 & 32.83 \\
(R) wrist, (R) thigh, (R) ear & 29.36 & 24.01 & 5.84 & 38.68 & 72.18 & 41.14 & 174.73 & 35.61 & 8.37 & 32.31 \\
\midrule
All placements (average) & 29.36 & 24.07 & 5.83 & 38.72 & 71.43 & 41.51 & 176.87 & 35.98 & 8.56 & 32.82 \\
All placements (std.) & $\pm$0.16 & $\pm$0.22 & $\pm$0.02 & $\pm$0.25 & $\pm$0.57 & $\pm$0.18 & $\pm$1.33 & $\pm$0.19 & $\pm$0.12 & $\pm$0.24 \\
\bottomrule
\end{tabular}
}
\label{tab:sensor_placement_humanml}
\end{table*}

\begin{table*}[t]
\caption{\small\textbf{Number of sensors on HumanML} (480 frames, 157k iteration checkpoint). Rows remove one sensor at a time, starting from the full 5pt setup; (L) and (R) denote the left and right side of the body. Motion error grows as sensors are removed, while text generation holds up far better: 3pt keeps 99\% of the 5pt CIDEr and 2pt keeps 91\%. With a single wrist sensor there is no reading near the pelvis, so the global trajectory drifts (MTE 289\,mm); a single thigh sensor avoids this.}
\centering
\resizebox{\linewidth}{!}{
\setlength{\tabcolsep}{2pt}
\begin{tabular}{@{} l l ccccc ccccc @{}}
\toprule
\multirow{2}{*}{\#IMU} & \multirow{2}{*}{IMU placement} & \multicolumn{5}{c}{Motion Metrics} & \multicolumn{5}{c}{Text Metrics} \\
\cmidrule(lr){3-7} \cmidrule(lr){8-12}
& & MPJPE $\downarrow$ & PA-MPJPE $\downarrow$ & MPJRE $\downarrow$ & MPJVE $\downarrow$ & MTE $\downarrow$
& ROUGE-L $\uparrow$ & CIDEr $\uparrow$ & BLEU@1 $\uparrow$ & BLEU@4 $\uparrow$ & BERT $\uparrow$ \\
\midrule
5pt & (L,R) wrist, (L,R) thigh, (L) ear & 23.60 & 17.66 & 4.99 & 30.59 & 63.55 & 42.15 & 180.09 & 36.63 & 8.78 & 33.71 \\
3pt & (L) wrist, (L) thigh, (L) ear     & 29.35 & 24.16 & 5.82 & 38.78 & 70.56 & 41.69 & 178.27 & 36.15 & 8.65 & 32.96 \\
\midrule
\multirow{2}{*}{2pt} & (L) wrist, (L) thigh & 38.01 & 30.57 & 7.50 & 60.04 & 83.01 & 40.35 & 164.17 & 34.57 & 7.18 & 30.80 \\
                     & (L) wrist, (R) thigh & 38.26 & 30.77 & 7.52 & 60.33 & 82.65 & 40.01 & 163.72 & 34.12 & 7.22 & 30.24 \\
\midrule
\multirow{2}{*}{1pt} & (L) wrist & 60.57 & 49.15 & 10.62 & 84.34 & 289.38 & 33.79 & 124.95 & 28.02 & 3.54 & 20.87 \\
                     & (L) thigh & 59.13 & 52.82 & 10.78 & 93.85 & 92.70  & 36.68 & 139.36 & 30.85 & 4.77 & 24.09 \\
\bottomrule
\end{tabular}
}
\label{tab:sensor_count_humanml}
\end{table*}

\subsection{Amount of Real-IMU Finetuning Data}

As shown in Tab.~\ref{tab:data_frac}, fine-tuning on only half of the real recordings already recovers most of the gain of full fine-tuning: on IMUPoser, 50\% of the data closes 94\% of the MPJPE gap between the zero-shot model and full fine-tuning (104.5 → 42.7 vs. 38.7 mm) and 97\% of the joint-rotation gap; on DIP-IMU it closes 68\% of the MPJPE gap and 89\% of the MPJRE gap.

\begin{table}[t]
\centering
\small
\caption{Effect of real-world fine-tuning data (480 frames, 3-point layout).
0\% is the model pretrained on synthetic-IMU without real-IMU fine-tuning.}
\label{tab:data_frac}
\begin{tabular}{llcccc}
\toprule
Dataset & Train data & MPJPE$\downarrow$ & PA-MPJPE$\downarrow$ & MPJRE$\downarrow$ & MPJVE$\downarrow$ \\
\midrule
\multirow{3}{*}{DIP-IMU}
 & 0\%   & 81.40 & 73.68 & 17.18 & 111.34 \\
 & 50\%  & 55.31 & 49.05 &  6.21 &  70.83 \\
 & 100\% & 43.24 & 33.91 &  4.91 &  52.30 \\
\midrule
\multirow{3}{*}{IMUPoser}
 & 0\%   & 104.46 & 81.20 & 19.99 & 180.73 \\
 & 50\%  &  42.66 & 36.56 &  6.14 &  53.75 \\
 & 100\% &  38.71 & 33.13 &  5.68 &  49.75 \\
\bottomrule
\end{tabular}
\end{table}

\subsection{Robustness to Body Shape Variations}
\label{sec:supp_body_shape}
\begin{figure}[t]
\centering
\begin{minipage}[c]{0.38\linewidth}
    \centering
    \includegraphics[width=\linewidth]{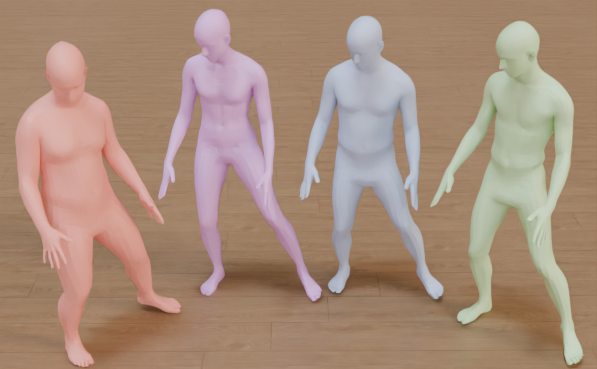}
    \caption{\small\textbf{Body shapes used in the robustness evaluation.}}
    \label{fig:different_beta_shape}
\end{minipage}
\hfill
\begin{minipage}[c]{0.58\linewidth}
    \centering
    \captionof{table}{\small\textbf{IMU-to-Motion under four body shapes.} Evaluated on HumanML on 480 frames with 5pt inputs. Our method is robust across various body shapes: the MPJPE of every variant stays within 0.3 mm of the default shape.}
    \setlength{\tabcolsep}{2.5pt}
    \resizebox{\linewidth}{!}{
\begin{tabular}{@{} l ccccc @{}}
\toprule
Body shape & MPJPE $\downarrow$ & PA-MPJPE $\downarrow$ & MPJRE $\downarrow$ & MPJVE $\downarrow$ \\
\midrule
Default & \textbf{14.69} & 11.32 & 3.06 & 19.31 \\
Short (1.526 m) & 14.99 & 11.54 & 3.08 & 19.57 \\
Tall (1.917 m) & 14.80 & 11.34 & 3.06 & 19.32 \\
Heavy (1.686 m) & 14.74 & \textbf{11.04} & \textbf{3.00} & \textbf{19.17} \\
\bottomrule
\end{tabular}
    }
    \label{tab:different_beta_shape}
\end{minipage}
\end{figure}

We map all motions to a default skeleton and do not explicitly model body-shape or gender variations. To assess robustness, we simulate IMU signals under three body-shape variations (short, tall, and heavy; Fig.~\ref{fig:different_beta_shape}) and directly evaluate our trained 5pt checkpoint on HumanML with 480-frame inputs. As shown in Tab.~\ref{tab:different_beta_shape}, the errors remain comparable to the default shape (MPJPE 14.69 vs.\ at most 14.99), indicating that our model is robust to body-shape variations. 

\section{More Qualitative Results}
\label{sec:supp_qualitative}

We provide additional qualitative results on (1) IMU-to-Motion comparisons with baselines, (2) text-conditioned object editing, (3) object sampling diversity, (4) object prediction on motion-only data, and (5) re-localization within a fixed 3D scene.

\subsection{IMU-to-Motion}
\label{sec:supp_qual_motion}
\begin{figure}[!ht]
\centering
\setlength{\tabcolsep}{2pt}
\resizebox{\linewidth}{!}{%
\begin{tabular}{@{}cccc@{}}
{\small GT} & {\small IMUPoser} & {\small MobilePoser} & {\small \textbf{Ours}} \\[2pt]
\multicolumn{4}{@{}l@{}}{\makebox[\dimexpr0.968\linewidth+12pt\relax]{\scriptsize (a) Side kicks \hfill $t=1.50, 1.90, 2.30$\,s}} \\[1pt]
\includegraphics[width=0.242\linewidth]{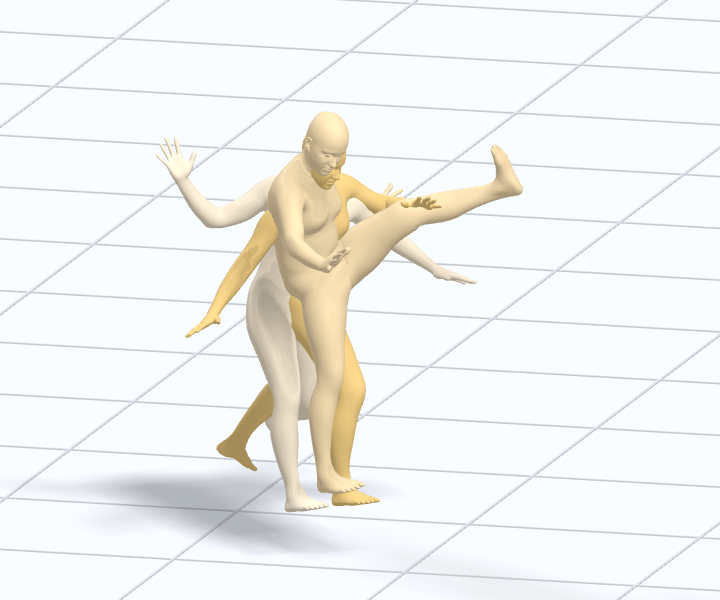} & \includegraphics[width=0.242\linewidth]{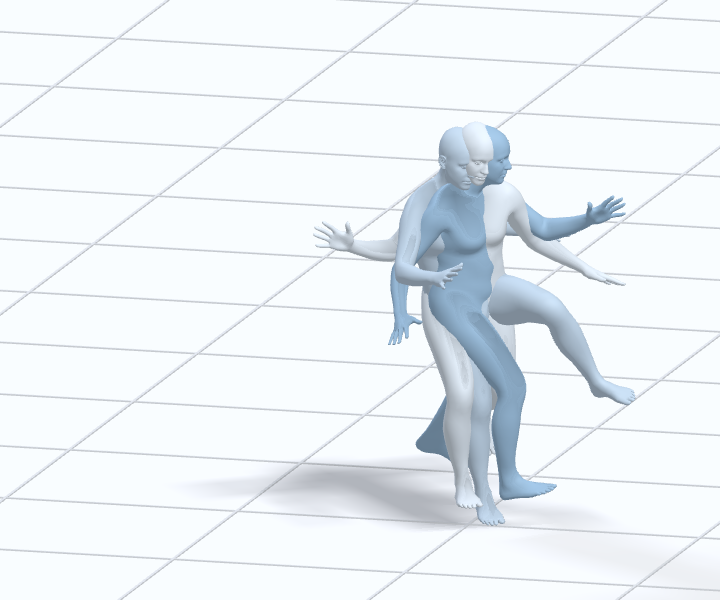} & \includegraphics[width=0.242\linewidth]{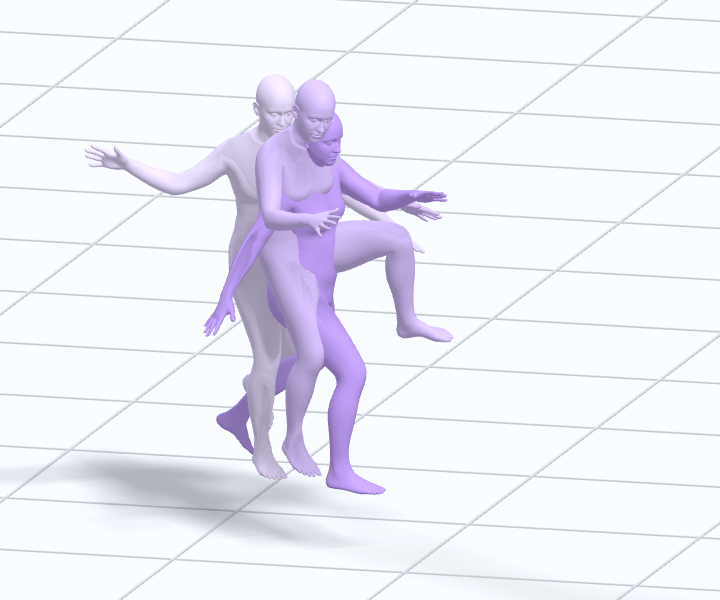} & \includegraphics[width=0.242\linewidth]{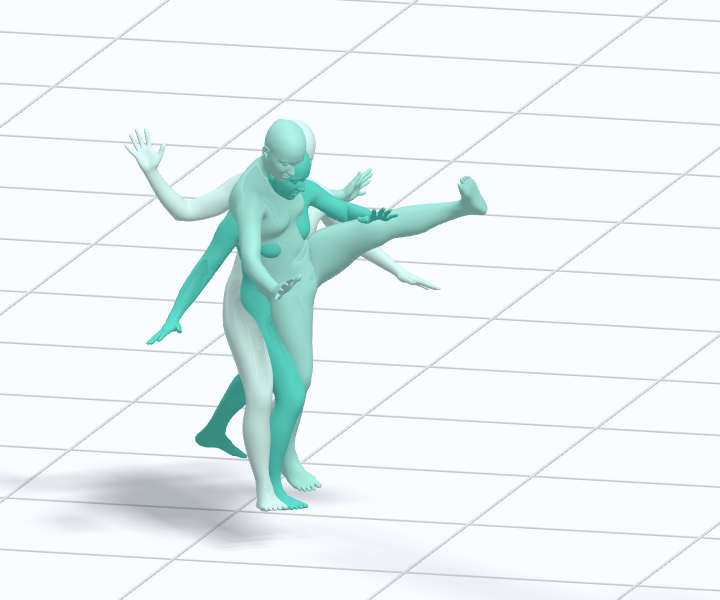} \\[4pt]
\multicolumn{4}{@{}l@{}}{\makebox[\dimexpr0.968\linewidth+12pt\relax]{\scriptsize (b) Backward bend into a bridge \hfill $t=6.10, 6.60, 7.10$\,s}} \\[1pt]
\includegraphics[width=0.242\linewidth]{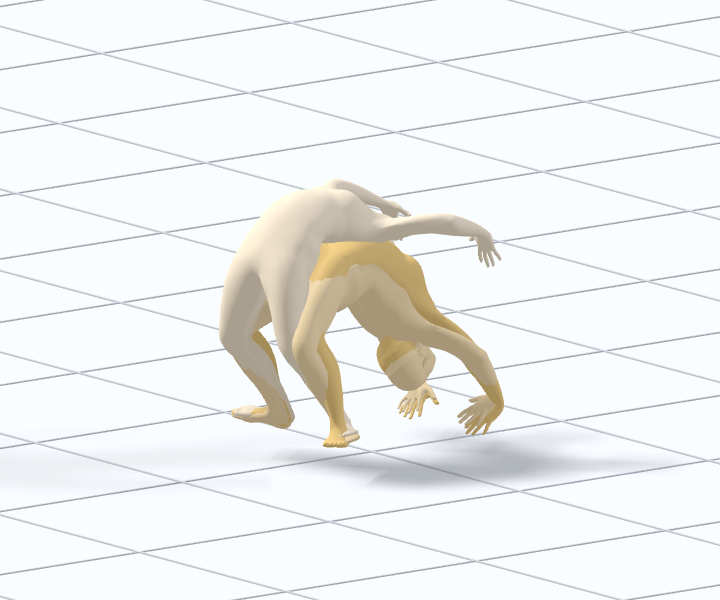} & \includegraphics[width=0.242\linewidth]{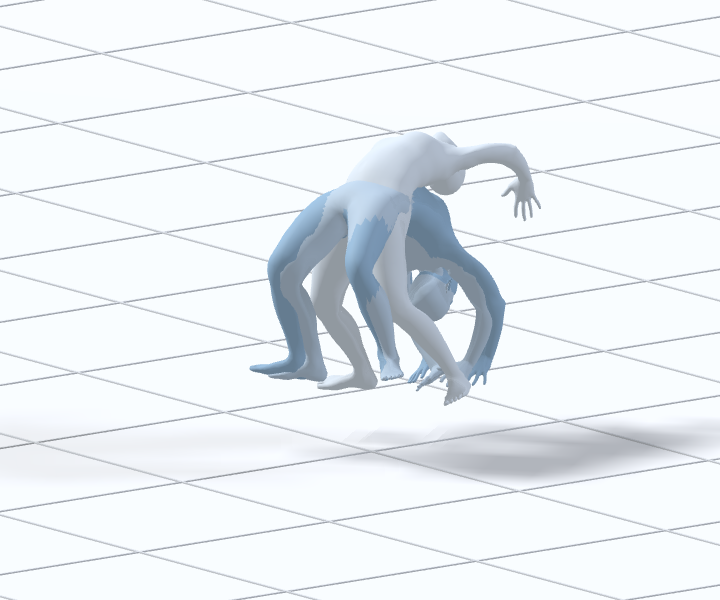} & \includegraphics[width=0.242\linewidth]{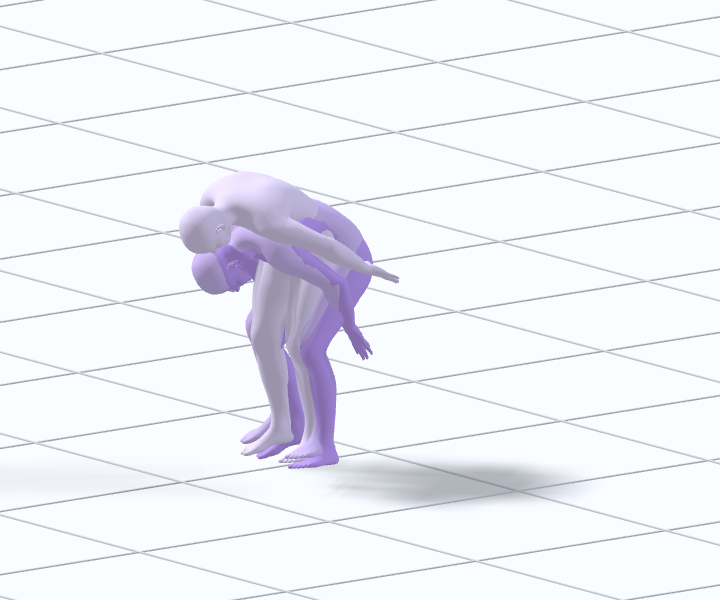} & \includegraphics[width=0.242\linewidth]{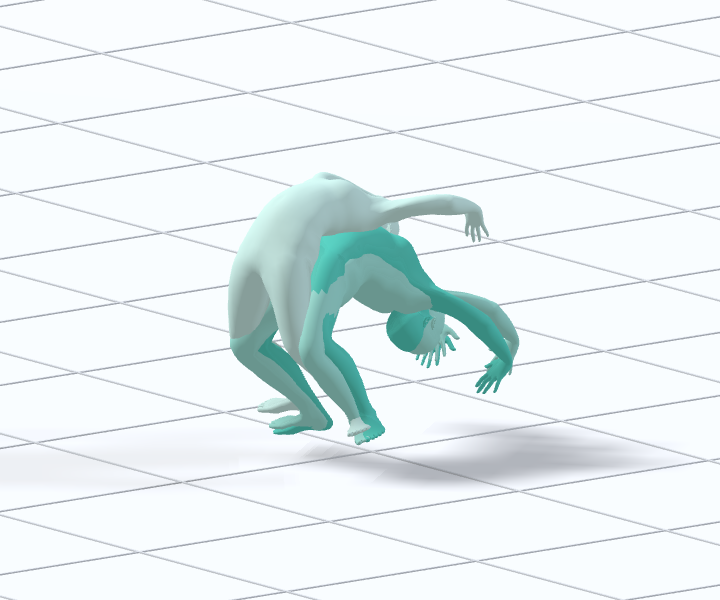} \\[4pt]
\multicolumn{4}{@{}l@{}}{\makebox[\dimexpr0.968\linewidth+12pt\relax]{\scriptsize (c) Floor dance \hfill $t=9.17, 9.57, 9.97$\,s}} \\[1pt]
\includegraphics[width=0.242\linewidth]{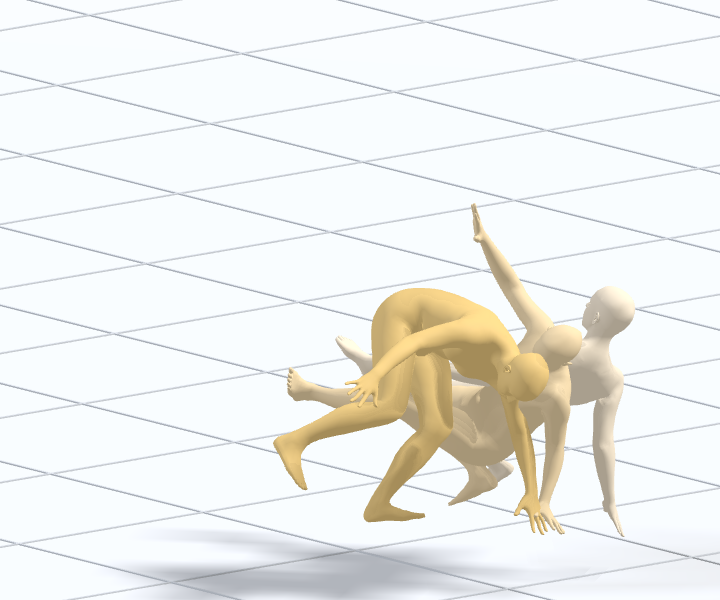} & \includegraphics[width=0.242\linewidth]{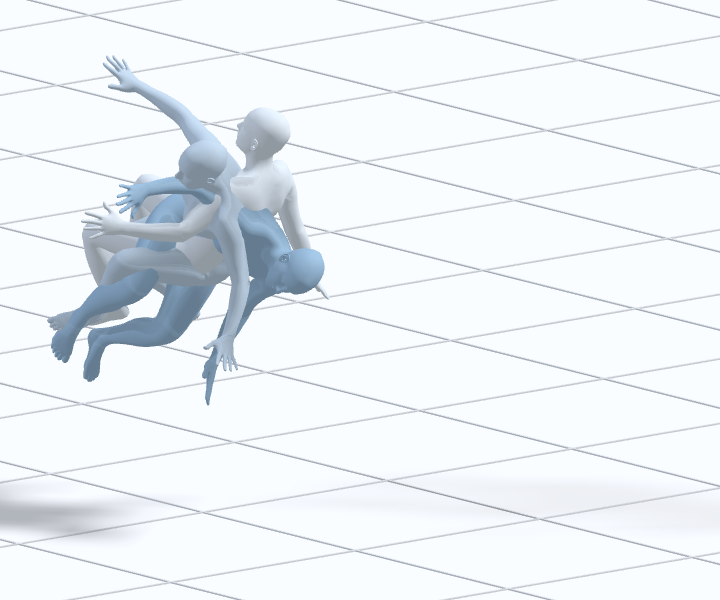} & \includegraphics[width=0.242\linewidth]{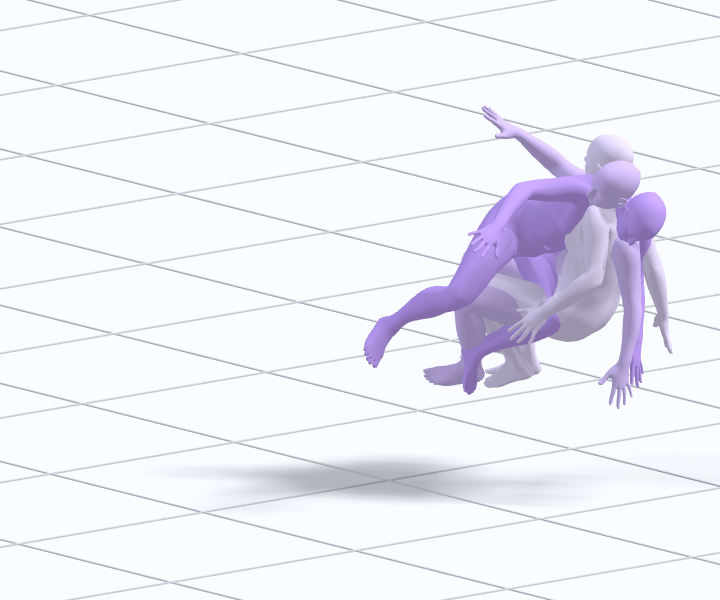} & \includegraphics[width=0.242\linewidth]{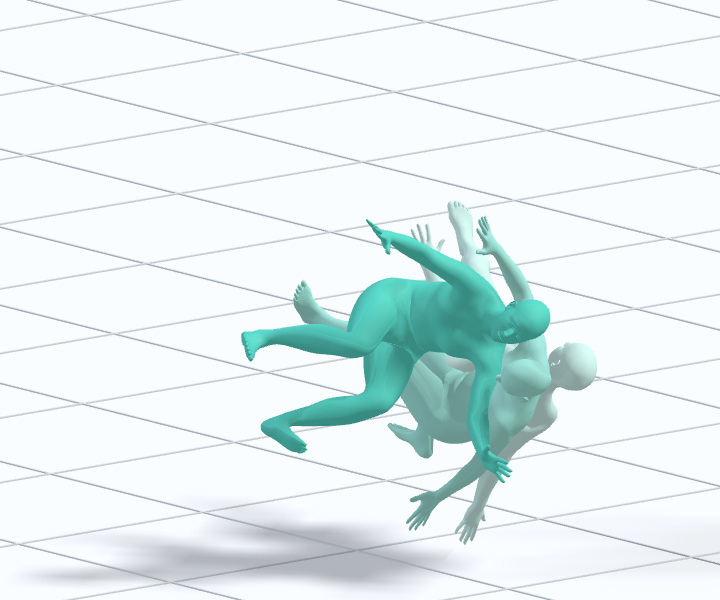} \\[4pt]
\multicolumn{4}{@{}l@{}}{\makebox[\dimexpr0.968\linewidth+12pt\relax]{\scriptsize (d) Moving with outstretched arms \hfill $t=2.17, 2.57, 2.97$\,s}} \\[1pt]
\includegraphics[width=0.242\linewidth]{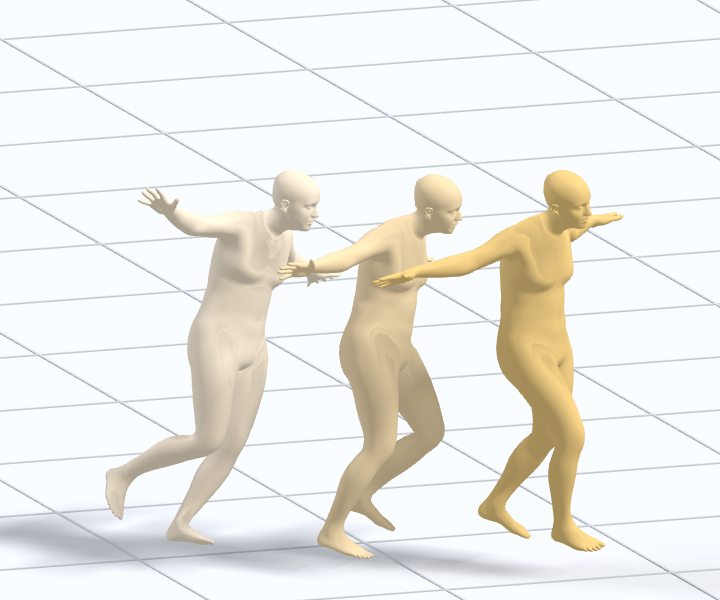} & \includegraphics[width=0.242\linewidth]{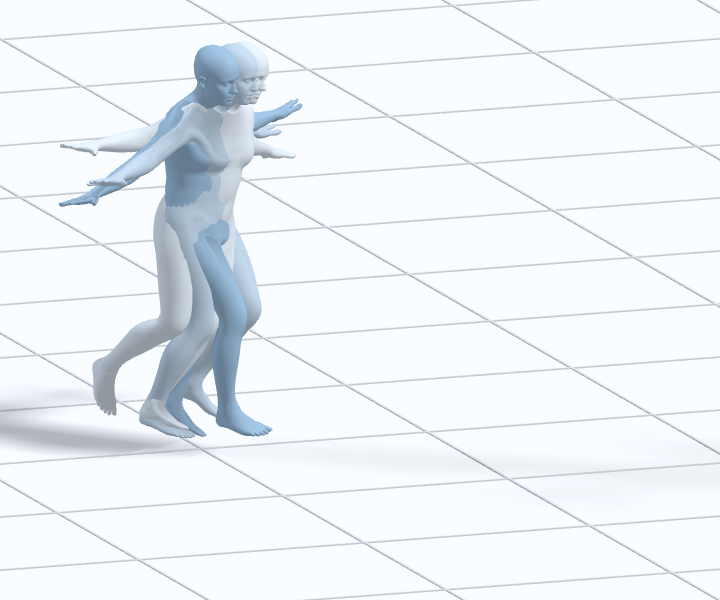} & \includegraphics[width=0.242\linewidth]{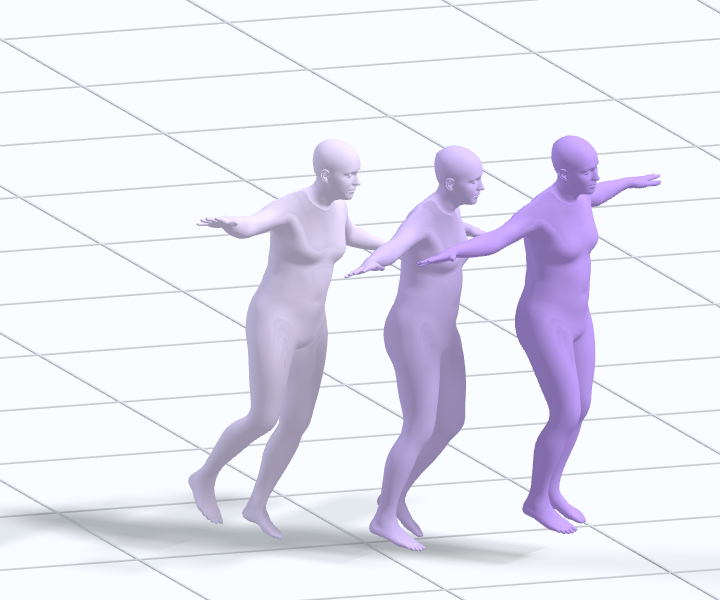} & \includegraphics[width=0.242\linewidth]{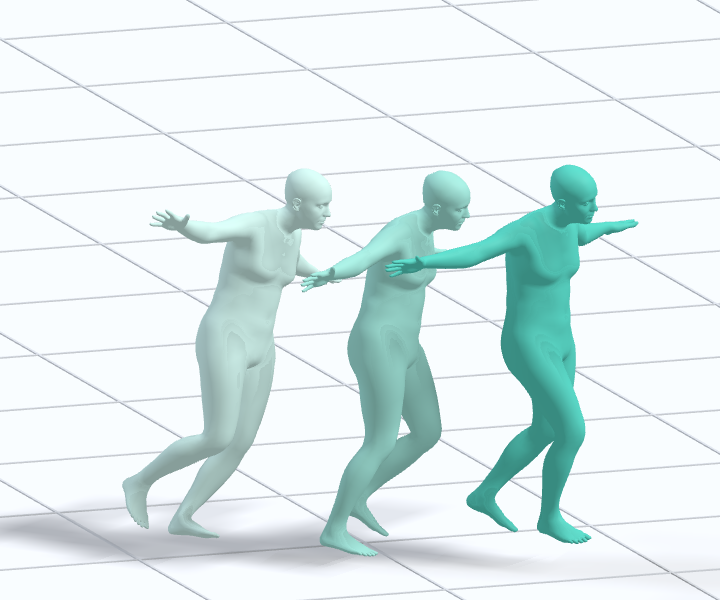} \\[4pt]
\end{tabular}}
\caption{\textbf{IMU-to-Motion on HumanML (5pt).} Each panel overlays three key poses sampled over 0.8--1.0\,s, from light to dark. All methods share the same frames and camera within each row, without per-method recentering. IMUPoser does not estimate global translation, so its poses remain near the starting position.}
\label{fig:supp_final_motion}
\end{figure}

Figure~\ref{fig:supp_final_motion} compares four large-amplitude motions on HumanML with the
5pt setup. In (a), our prediction reproduces the high, extended side kick, whereas IMUPoser and
MobilePoser only raise the knee. In (b), our model recovers the backward bend into a bridge;
MobilePoser instead bends forward, and IMUPoser's poses slide sideways. Row (d) mainly
illustrates global translation: both MobilePoser and our model follow the forward displacement,
while IMUPoser does not estimate global translation by design. Floor-level motion (c) remains
challenging for all methods, including ours: the predicted bodies float above the floor,
since body height is not directly observed by the IMUs.

\subsection{Text-Conditioned Object Editing}
\label{sec:supp_qual_editing}
\begin{figure}[!ht]
\centering
\setlength{\tabcolsep}{2pt}
\resizebox{\linewidth}{!}{%
\begin{tabular}{@{}cccc@{}}
{\small Original: ``chair''} & {\small ``chair'' $\rightarrow$ ``box''} & {\small ``chair'' $\rightarrow$ ``table''} & {\small ``chair'' $\rightarrow$ ``trash can''} \\[2pt]
\multicolumn{4}{@{}l}{\scriptsize $t=7.20$\,s} \\[1pt]
\includegraphics[width=0.242\linewidth]{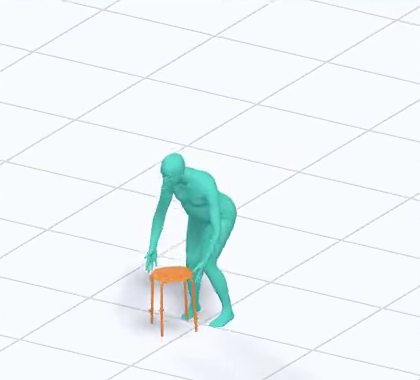} & \includegraphics[width=0.242\linewidth]{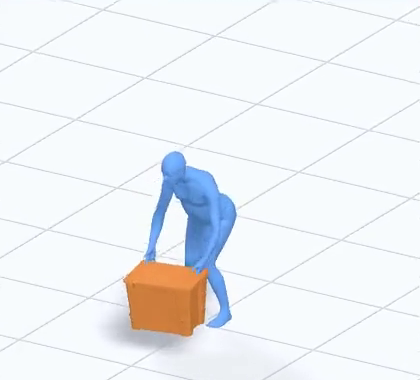} & \includegraphics[width=0.242\linewidth]{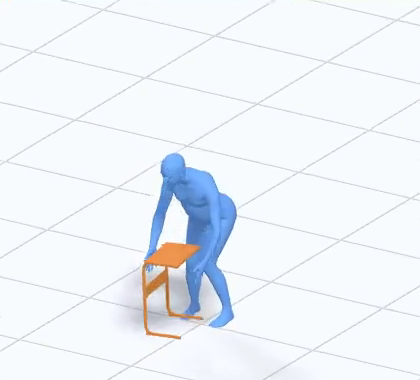} & \includegraphics[width=0.242\linewidth]{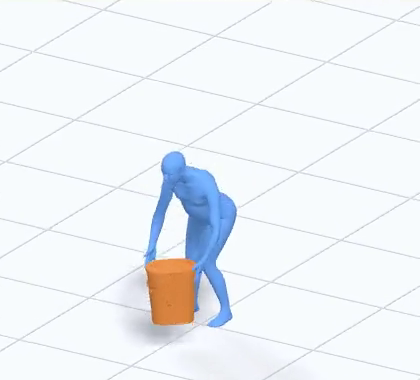} \\[4pt]
\multicolumn{4}{@{}l}{\scriptsize $t=11.20$\,s} \\[1pt]
\includegraphics[width=0.242\linewidth]{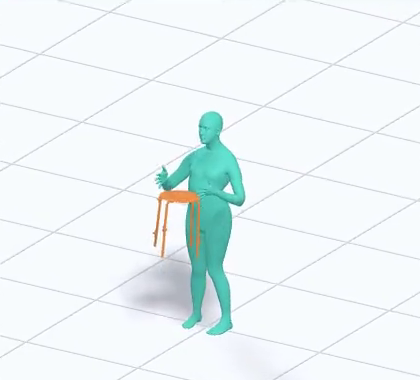} & \includegraphics[width=0.242\linewidth]{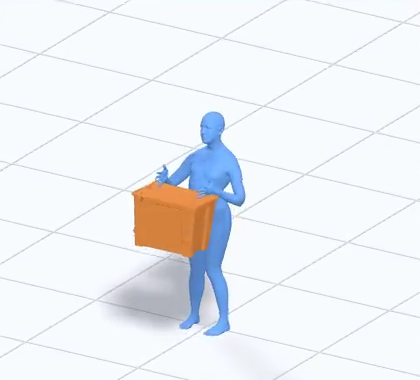} & \includegraphics[width=0.242\linewidth]{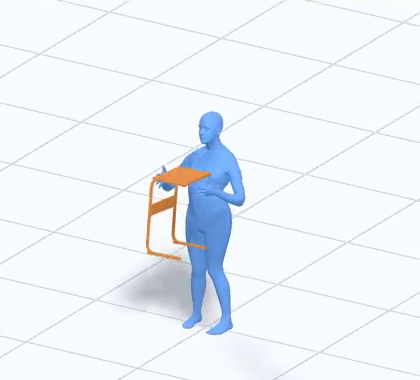} & \includegraphics[width=0.242\linewidth]{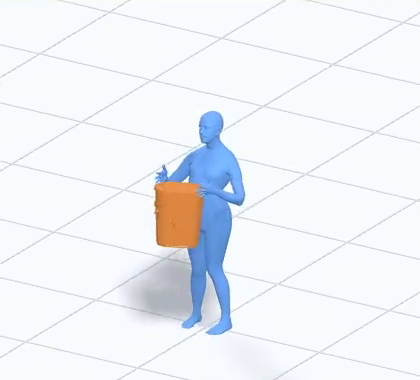} \\[4pt]
\multicolumn{4}{@{}l}{\scriptsize $t=14.40$\,s} \\[1pt]
\includegraphics[width=0.242\linewidth]{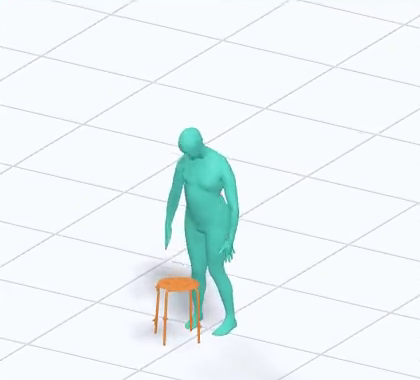} & \includegraphics[width=0.242\linewidth]{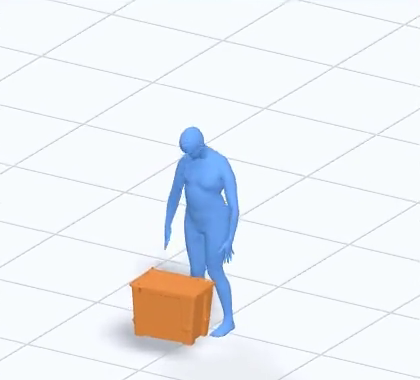} & \includegraphics[width=0.242\linewidth]{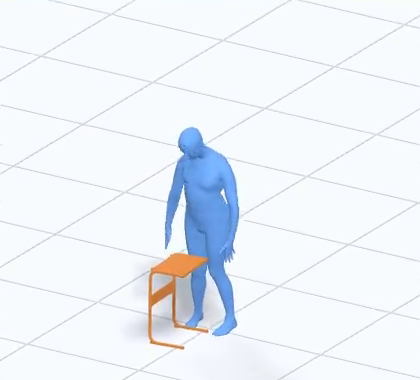} & \includegraphics[width=0.242\linewidth]{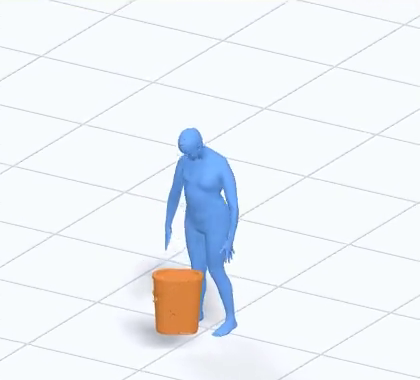} \\[4pt]
\end{tabular}}
\caption{\textbf{Text-conditioned object editing on HiPhi.}  The first column shows our original
prediction (green); the other columns (blue) show the objects re-decoded after replacing
``chair'' in the generated caption with the listed object names.
The human motion is identical across columns. Rows show picking up, carrying, and putting down
at the same times.}
\label{fig:supp_final_editing}
\end{figure}

Since the output sequence is ordered as IMU$\rightarrow$Motion$\rightarrow$Text$\rightarrow$Object,
editing the caption changes only the objects decoded after it, while the predicted motion is
unaffected. We replace the object noun in the generated caption and re-decode the object tokens. In Fig.~\ref{fig:supp_final_editing}, the original chair is replaced with a box, table, or trash can across the lifting sequence. This illustrates how text can control the inferred interaction context for the same motion. The edited objects follow the broad lifting pattern, although precise hand--object contact is not guaranteed.

\subsection{Object Sampling Diversity}
\label{sec:supp_qual_diversity}
\begin{figure}[!ht]
\centering
\setlength{\tabcolsep}{2pt}
\resizebox{\linewidth}{!}{%
\begin{tabular}{@{}cccc@{}}
{\small GT} & {\small Sample 1: table} & {\small Sample 2: box} & {\small Sample 3: box} \\[2pt]
\includegraphics[width=0.242\linewidth]{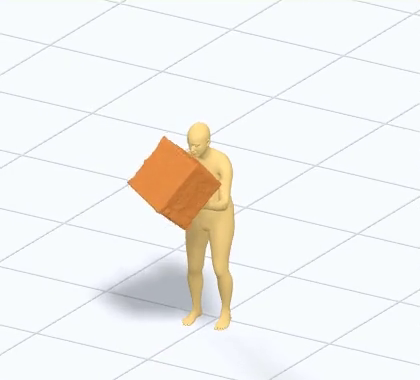} & \includegraphics[width=0.242\linewidth]{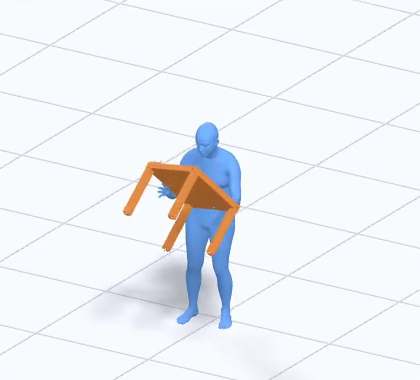} & \includegraphics[width=0.242\linewidth]{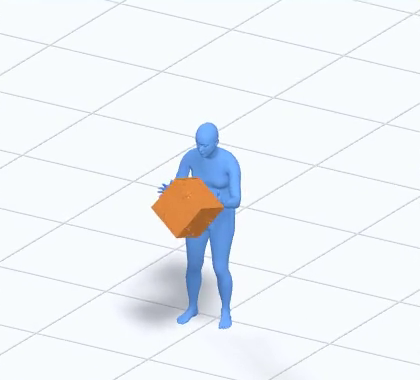} & \includegraphics[width=0.242\linewidth]{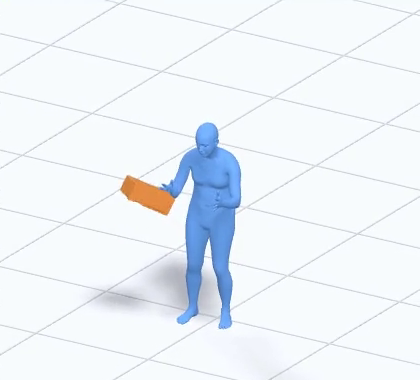} \\[4pt]
\end{tabular}}
\caption{\textbf{Object sampling diversity on OMOMO.} GT (yellow; a box) and three samples
(blue) from the same IMU input at $t=3.47$\,s. Sampling yields different object categories and
instances that are each compatible with the carrying motion.}
\label{fig:supp_final_diversity}
\end{figure}

IMU measurements constrain body motion but do not directly reveal the identity or geometry of
the carried object. We therefore sample the object tokens while keeping the predicted motion and caption fixed. In Fig.~\ref{fig:supp_final_diversity},
the same two-handed carrying motion yields a small table and two box instances of different
sizes. Carrying a small table against the chest is a common OMOMO interaction and is hard to
distinguish from carrying a box by body motion alone, so sampling exposes this ambiguity rather
than resolving it.

\subsection{Object Prediction on Motion-Only Data}
\label{sec:supp_qual_ood}
\begin{figure}[!ht]
\centering
\setlength{\tabcolsep}{2pt}
\resizebox{\linewidth}{!}{%
\begin{tabular}{@{}cc@{\hspace{12pt}}cc@{}}
\multicolumn{2}{c}{\small (a) Dribbling $\rightarrow$ ball} & \multicolumn{2}{c}{\small (b) Leaning $\rightarrow$ table} \\
\cmidrule(lr){1-2}\cmidrule(lr){3-4}
{\footnotesize GT motion} & {\footnotesize Ours} & {\footnotesize GT motion} & {\footnotesize Ours} \\[2pt]
\multicolumn{2}{c}{\scriptsize $t=0.80$\,s} & \multicolumn{2}{c}{\scriptsize $t=1.87$\,s} \\[1pt]
\includegraphics[width=0.242\linewidth]{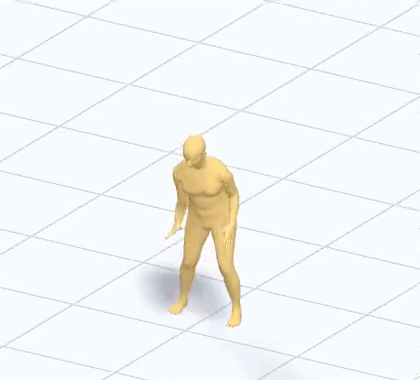} & \includegraphics[width=0.242\linewidth]{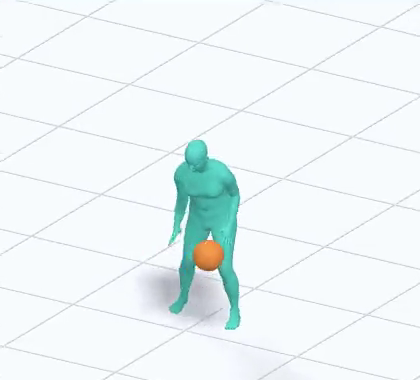} & \includegraphics[width=0.242\linewidth]{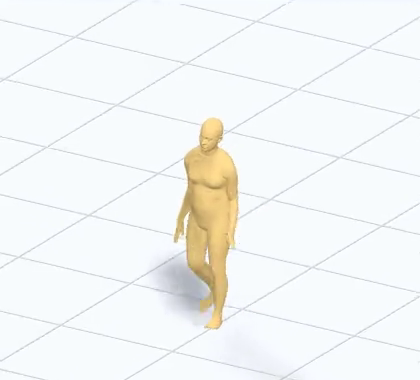} & \includegraphics[width=0.242\linewidth]{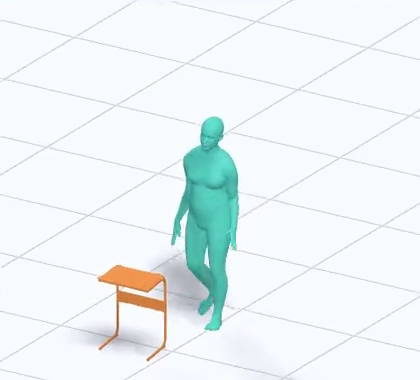} \\[4pt]
\multicolumn{2}{c}{\scriptsize $t=1.80$\,s} & \multicolumn{2}{c}{\scriptsize $t=4.20$\,s} \\[1pt]
\includegraphics[width=0.242\linewidth]{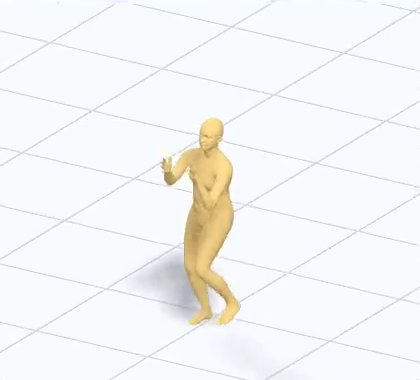} & \includegraphics[width=0.242\linewidth]{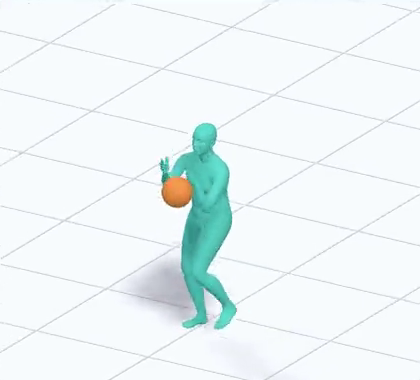} & \includegraphics[width=0.242\linewidth]{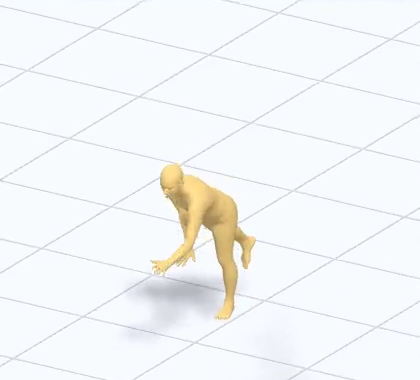} & \includegraphics[width=0.242\linewidth]{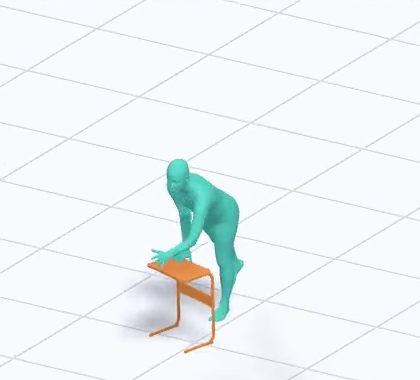} \\[4pt]
\multicolumn{2}{c}{\scriptsize $t=2.80$\,s} & \multicolumn{2}{c}{\scriptsize $t=6.47$\,s} \\[1pt]
\includegraphics[width=0.242\linewidth]{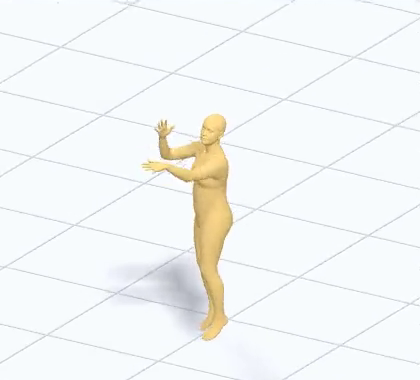} & \includegraphics[width=0.242\linewidth]{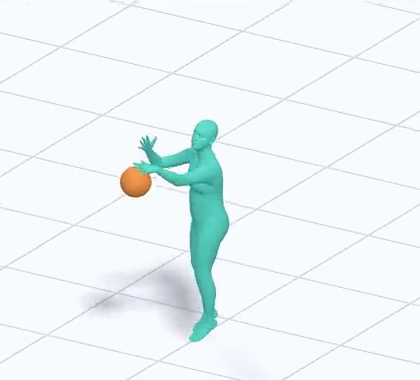} & \includegraphics[width=0.242\linewidth]{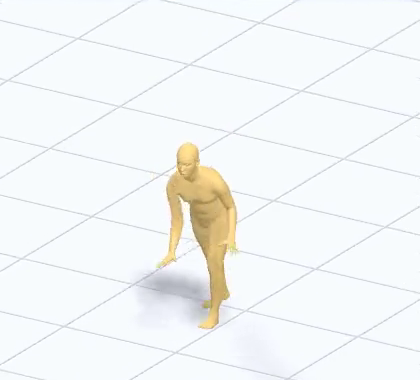} & \includegraphics[width=0.242\linewidth]{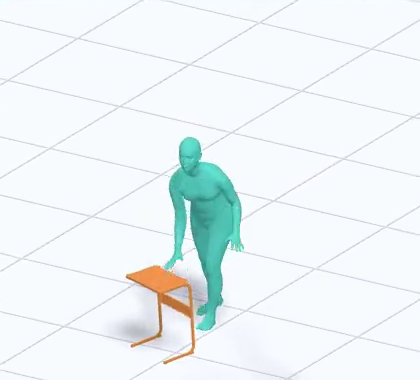} \\[4pt]
\end{tabular}}
\caption{\textbf{Object prediction on HumanML sequences without object annotations.} GT motion (yellow) and our predicted human and object (green) at the same times.
(a) For a dribbling motion, the predicted ball moves with the hands over time.
(b) For a leaning motion, the predicted table appears beneath the hands.
No object ground truth is available; see Sec.~\ref{sec:supp_qual_ood}.}
\label{fig:supp_final_ood}
\end{figure}

Large motion corpora such as HumanML lack object annotations, while human--object interaction
datasets are much smaller. We therefore ask whether the motion--object associations learned from
interaction data transfer to motion-only sequences. We apply our model finetuned on HUMOTO to
HumanML. By default, it predicts no object for most sequences, as most HumanML motions involve no
object interaction; to probe which object the model associates with a motion, we suppress the
end-of-object token until one non-ground object is generated. In Fig.~\ref{fig:supp_final_ood},
the model predicts a ball for a dribbling motion and a table for a leaning motion. Beyond the
category, the predicted placement follows the motion over time. Note that these examples are meant to show cross-dataset transfer and remain qualitative.

\subsection{Re-localization in a Known Scene}
\label{sec:supp_qual_reloc}
\begin{figure}[!ht]
    \centering
    \resizebox{\linewidth}{!}{%
    \begin{tabular}{ccccc}
        \includegraphics[width=0.19\linewidth]{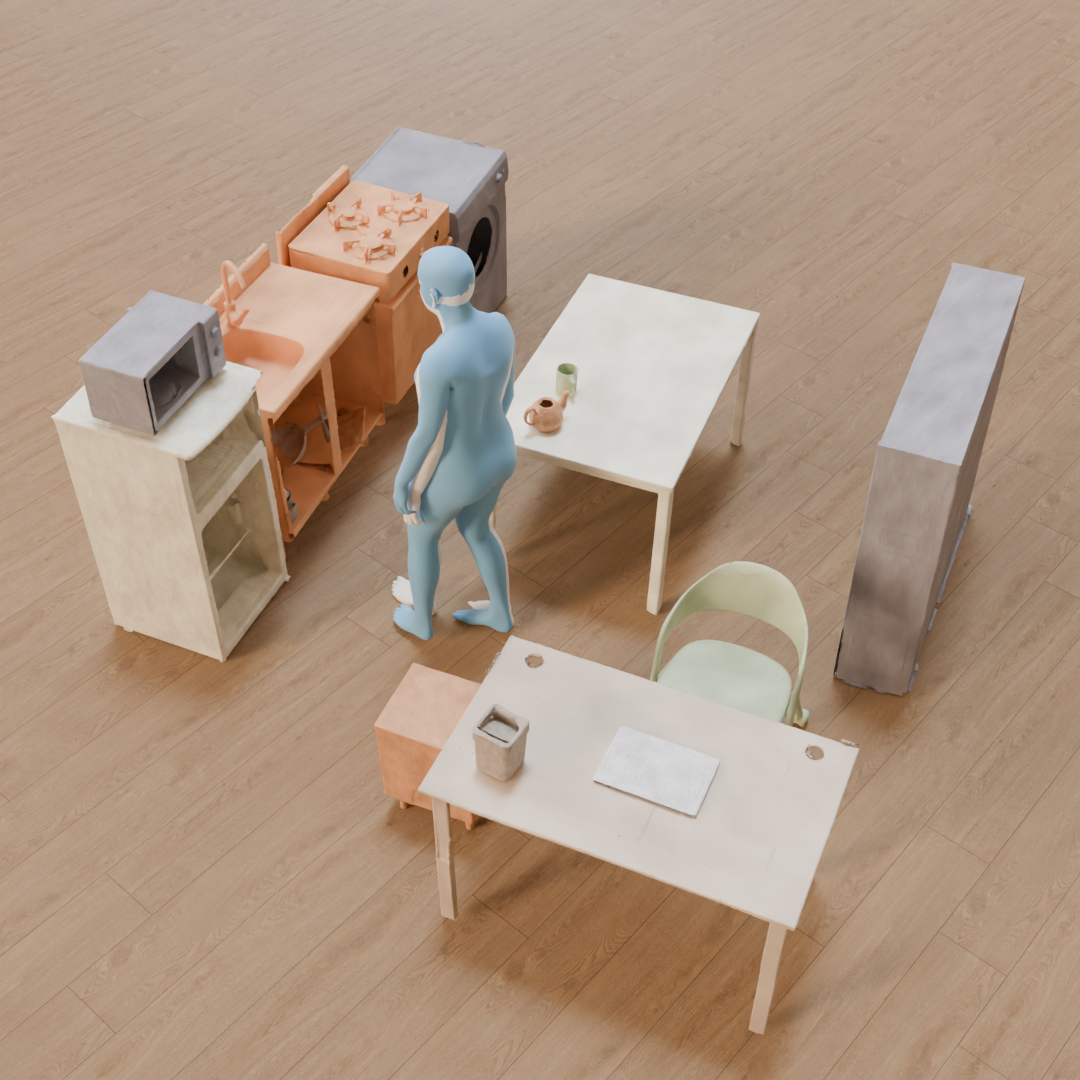} &
        \includegraphics[width=0.19\linewidth]{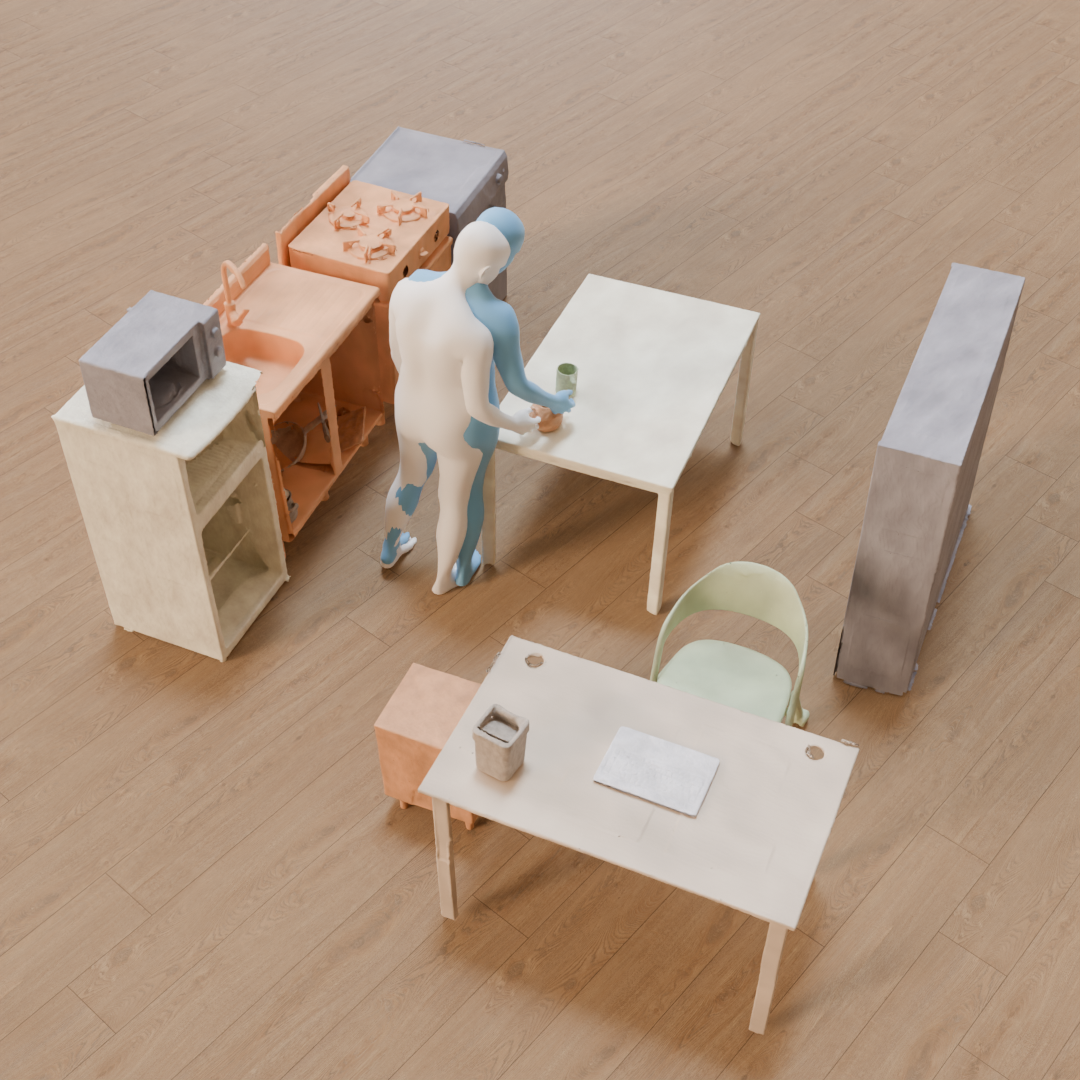} &
        \includegraphics[width=0.19\linewidth]{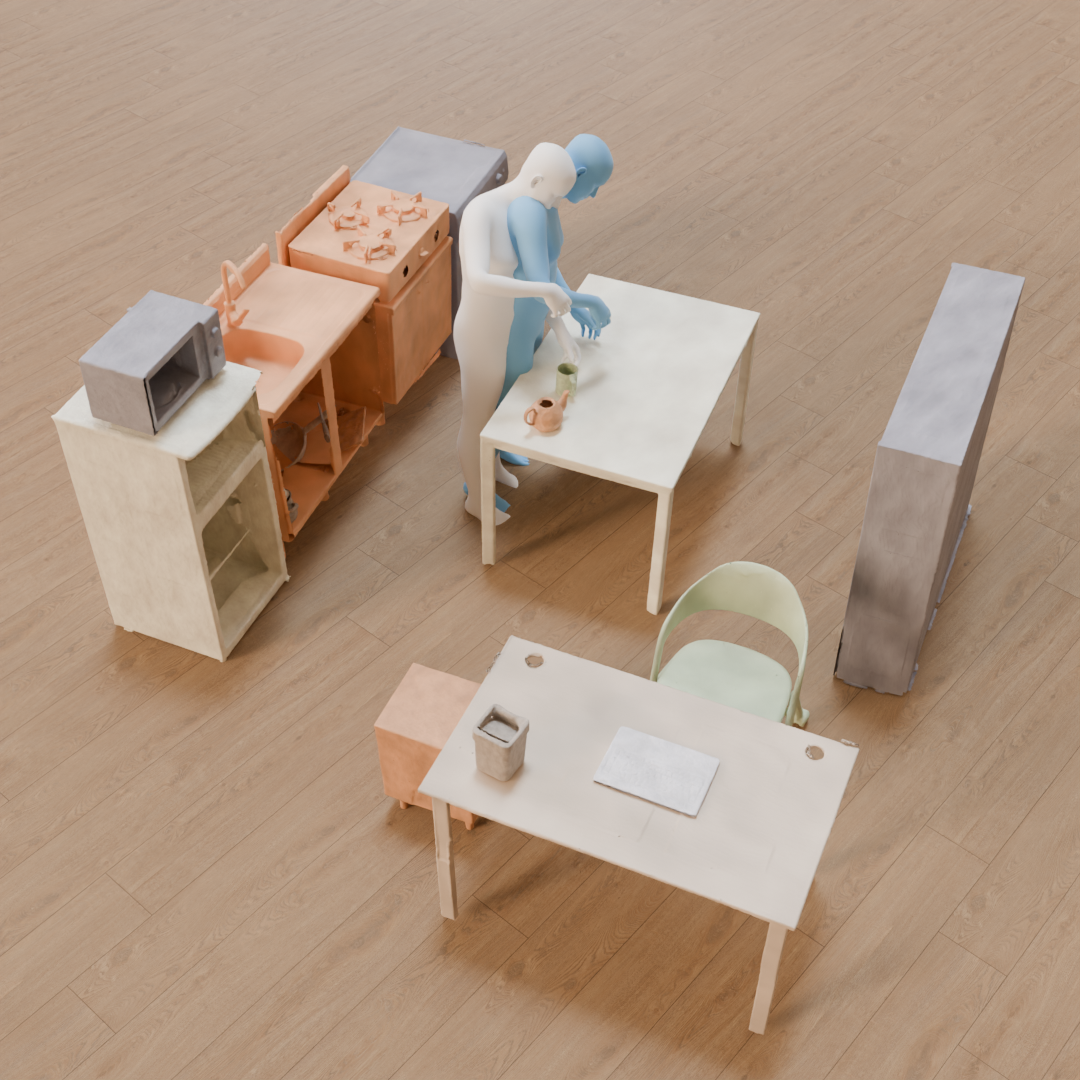} &
        \includegraphics[width=0.19\linewidth]{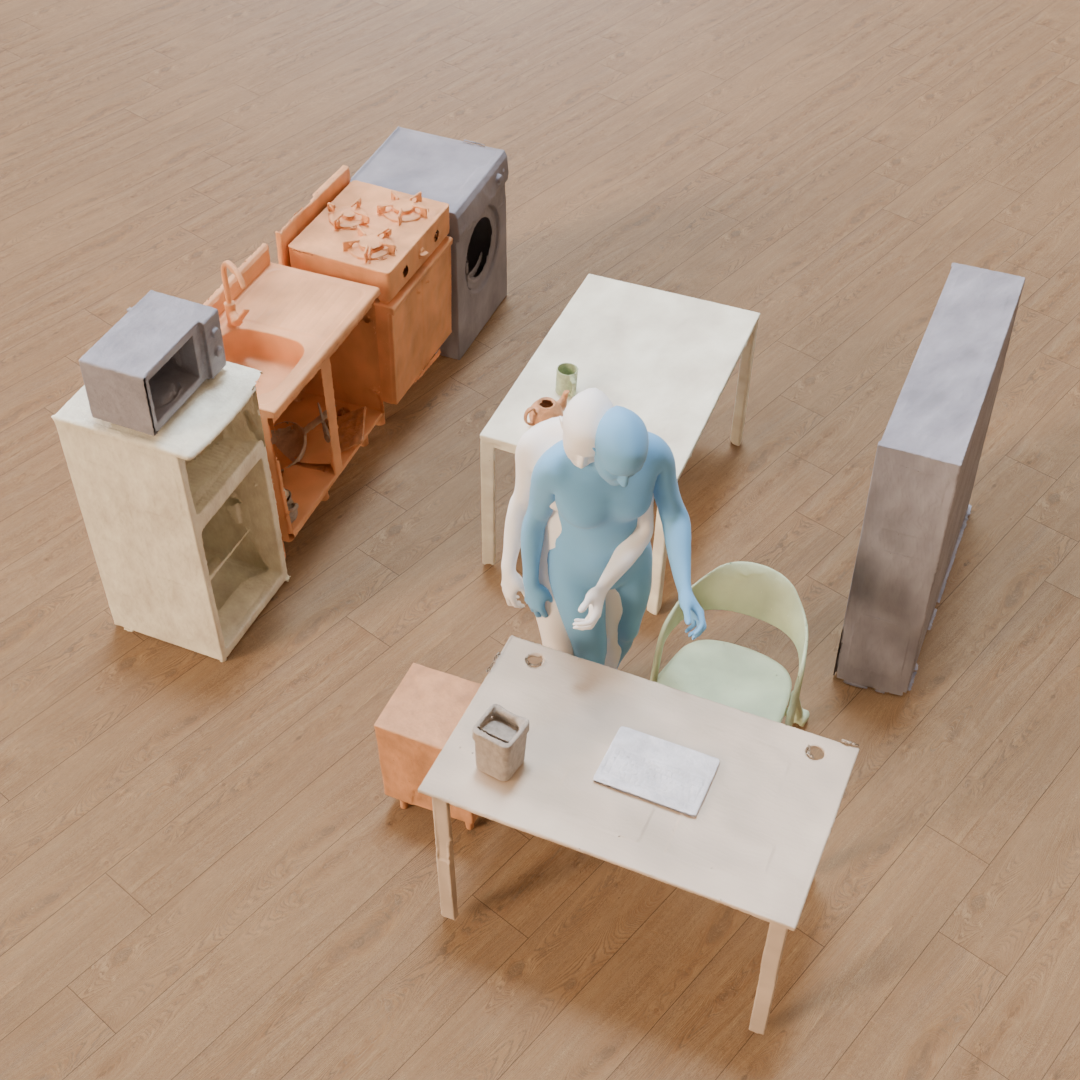} &
        \includegraphics[width=0.19\linewidth]{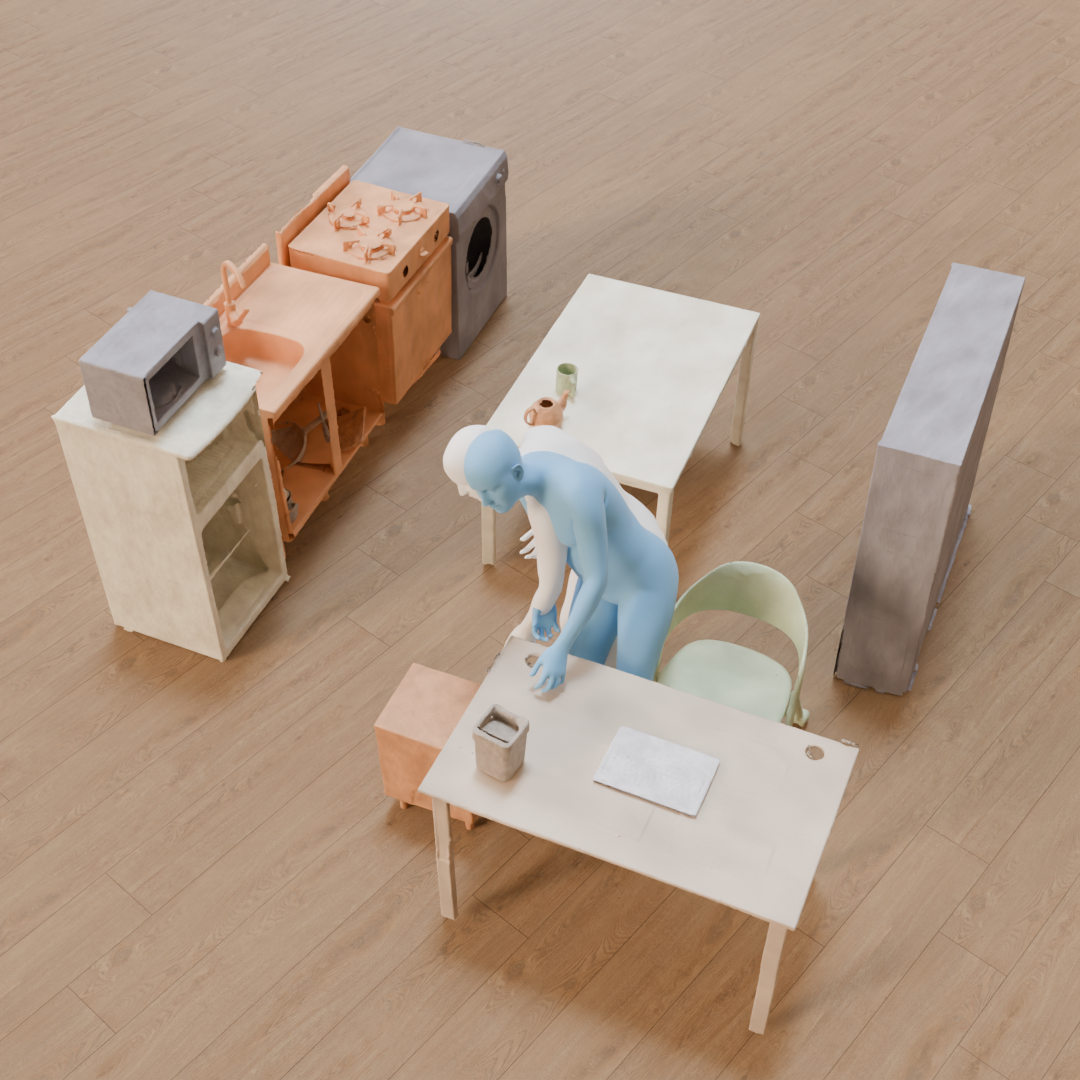} 
        \vspace{-3pt} \\
        {\footnotesize Frame 0} &
        {\footnotesize Frame 132} &
        {\footnotesize Frame 227} &
        {\footnotesize Frame 617} &
        {\footnotesize Frame 955} \\
    \end{tabular}%
    }
\caption{\textbf{Re-localization on ParaHome~\citep{kim2025parahome}.} White: ground truth; blue: our prediction. 
}
\label{fig:qual_reloc_parahome}
\end{figure}

IMU signals measure only relative motion and carry no absolute position. When finetuned on
sequences from a single fixed scene of ParaHome (Sec.~\ref{sec:supp_data_reloc}), our model
additionally predicts the person's initial global position and heading from IMU input alone,
likely by learning a scene-specific prior that links activities to plausible locations.
This capability is scene-specific and requires
in-scene training data; it does not transfer to unseen environments.
An example of re-localization is shown in Fig.~\ref{fig:qual_reloc_parahome}.

\FloatBarrier
\section{Implementation Details}
\label{sec:supp_implementation_details}

\subsection{Training}

\noindent{\textbf{Data augmentation.}}
During training, we apply three augmentation strategies to improve robustness. First, we randomly sample a plausible IMU sensor configuration and mask unused sensors to simulate real-world scenarios. Second, each IMU–motion sequence is randomly cropped and re-aligned to the first frame to enhance temporal robustness. Third, we randomly drop text descriptions with probability $p=0.3$ to accommodate datasets without text annotations. Since our sequence follows the order IMU$\to$Motion$\to$Text$\to$Object, masking text during training also prevents erroneous tokens from propagating to the downstream 3D object prediction.

\noindent{\textbf{Training objective.}}
All three outputs are trained jointly with a single sequence layout
$[\mathbf{S}, \mathbf{q}_{1:N}, \mathbf{D}, \mathcal{O}]$,
where $\mathbf{S}$, $\mathbf{D}$, and $\mathcal{O}$ denote the embedded IMU signals, the motion description, and the object list (Sec.~\ref{sec:prob_form}).
\emph{Motion.} Instead of teacher-forcing ground-truth motion tokens, the motion span
consists of $N=T/d_\mathrm{win}$ learnable queries $\mathbf{q}_n=\mathbf{e}_{q}+\mathrm{PE}(n)$, where $\mathbf{e}_q$ is a shared learnable embedding and $\mathrm{PE}(n)$ is the positional encoding of window $n$, one per
$d_\mathrm{win}$-frame window, that attend causally to the IMU prefix and bidirectionally to each other,
so the full motion is decoded in one parallel pass and the training and inference inputs
of this span are identical. Following TOTEM-style tokenization~\citep{talukder2024totem}, each of the nine channels of the root
translation and orientation increments (3+6) in a window is represented by its own normalized-shape, mean, and
standard-deviation tokens ($\mathbf{e}_1$, $\mathbf{e}_2$, and $\mathbf{e}_3$ in Sec.~\ref{sec:tokenize}), \ie, 27 tokens per window, all predicted with cross-entropy.
The shape token indexes a 4,096-entry codebook of a time-series VQ-VAE, which is pretrained on all available motion trajectories and kept frozen.
The mean and standard deviation are each quantized into 1,024 bins over $[-3, 3]$ and $[0, 0.5]$, respectively, with bin edges uniformly spaced in signed square-root space so that values near zero are resolved more finely.
The 21 local joint rotations $\boldsymbol\theta$ (6D) are regressed directly:
\begin{equation}
\mathcal{L}_{\mathrm{motion}} = \frac{1}{3}\left(\mathcal{L}^{\mu}_{\mathrm{CE}} + \mathcal{L}^{\sigma}_{\mathrm{CE}} + \mathcal{L}^{s}_{\mathrm{CE}}\right) + \lambda_{\theta}\,\mathrm{SmoothL1}_{\beta=0.02}(\hat{\boldsymbol\theta},\boldsymbol\theta).
\end{equation}
\emph{Text and objects.} The caption and the object category tokens are generated
autoregressively and trained with teacher forcing ($\mathcal{L}_{\mathrm{text}}$, $\mathcal{L}_{\mathrm{cat}}$). Objects are represented at two levels. The category token, supervised by $\mathcal{L}_{\mathrm{cat}}$, is a cross-entropy over the $C$ object categories and shares its vocabulary with the caption (\eg, ``chair''). The instance, supervised by $\mathcal{L}_{\mathrm{id}}$, selects a specific object mesh from a closed set (\eg, a dining chair or a low stool); since instances within a category differ in size and shape, this choice determines how the object fits the motion and which extent is used for its bounding box.
Since they attend to the motion-query hidden states rather
than to ground-truth motion, the language and object heads are conditioned on the model's
own motion estimate both at training and at test time. For each object, the model
predicts its first-frame pose $\mathbf{P}_i$, supervised with $\mathcal{L}_{\mathrm{anc}}$, and its asset
identity, supervised with $\mathcal{L}_{\mathrm{id}}$; the ground-truth pose is embedded and
fed as the next input token (the prediction is fed at inference). A temporal track decoder
reads the whole motion-query sequence together with the root trajectory (ground truth
during training, the decoded one at inference) and predicts per-frame object poses $\mathbf{P}_i^t$ in the
body frame, supervised on moving frames by
$\mathcal{L}_{\mathrm{track}}=\lambda_{\mathrm{tr}}(\mathcal{L}_{\mathrm{pos}}+2\mathcal{L}_{\mathrm{vel}}+2\mathcal{L}_{\mathrm{acc}})$,
and a per-frame moving state $m_i^t$ by a binary cross-entropy $\mathcal{L}_{\mathrm{state}}$.
Object losses are averaged over the objects that carry supervision. The overall objective is
\begin{equation}
\mathcal{L} = \mathcal{L}_{\mathrm{motion}} + \lambda_{\mathrm{text}}\mathcal{L}_{\mathrm{text}} + \mathcal{L}_{\mathrm{cat}} + \lambda_{\mathrm{anc}}\mathcal{L}_{\mathrm{anc}} + \lambda_{\mathrm{id}}\mathcal{L}_{\mathrm{id}} + \mathcal{L}_{\mathrm{track}} + \lambda_{\mathrm{st}}\mathcal{L}_{\mathrm{state}},
\end{equation}
with $\lambda_{\theta}=15.6$, $\lambda_{\mathrm{text}}=\lambda_{\mathrm{anc}}=\lambda_{\mathrm{id}}=1$, $\lambda_{\mathrm{tr}}=50$ and $\lambda_{\mathrm{st}}=5$.
We first train only the modality adapters and output heads for 1k steps with the
backbone frozen, and then train all parameters for 500k steps with AdamW~\citep{kingma2014adam}
(learning rate $10^{-4}$, 1k warm-up steps, constant afterwards) on randomly sampled
sensor layouts and crop lengths of 0.8--16\,s; we report the EMA weights (decay 0.999).

\noindent{\textbf{Human--object interaction finetuning.}}
For Stage-2 finetuning on the human--object interaction datasets (Sec.~\ref{sec:supp_data_hoi}), we initialize from the Stage-1 pretrained checkpoint and finetune a separate model on each dataset. We finetune for 50k steps on HUMOTO and OMOMO, and for 100k steps on HiPhi.

\subsection{Inference}

\noindent{\textbf{Sampling stop condition.}}
Given an IMU input sequence of length $T$, the model outputs human motion of the same length $T$. For text and object generation, decoding proceeds autoregressively until the model emits the special end tokens $\texttt{<EOT>}$ and $\texttt{<EOOBJ>}$, where $\texttt{<EOT>}$ and $\texttt{<EOOBJ>}$ are the stop tokens of text and object respectively.

\noindent{\textbf{Shifted window with pose averaging.}}
Given an IMU input of length $T$, we perform two overlapping inference passes: the first produces prediction $m_1$ over the full sequence, and the second produces a shifted prediction $m_2$ by offsetting the input by two frames. The final motion is obtained by averaging over the overlapping region as $m_{\mathrm{final}}=(m_1+m_2)/2$, which yields smoother and more accurate results in local coordinates.

\section{Dataset Details}
\label{sec:supp_dataset_details}

\subsection{Data Splits}
\label{sec:supp_data_splits}
All baselines and our model are trained on the same training split and evaluated on the same test clips. For the motion corpus, we use the official MotionMillion split. LINGO is split 80/10/10, and HiPhi, OMOMO, and HUMOTO are split separately from the motion corpus. Because MotionMillion assigns mirrored copies to splits independently, some HumanML test clips have a mirrored counterpart in the training split. This affects all methods equally; LINGO and the HOI test sets are free of such overlap. Tab.~\ref{tab:dataset_stats} summarizes the statistics of all datasets.
\begin{table}[t]
\caption{\small\textbf{Dataset statistics.} Number of sequences in the training and evaluation splits, and total duration over all splits at 30\,fps. $^\dagger$Including mirrored copies. $^\ddagger$Evaluated on 2,000 HumanML test clips without mirrored copies and on the 1,945 LINGO test clips.}
\label{tab:dataset_stats}
\centering
\small
\setlength{\tabcolsep}{6pt}
\begin{tabular}{@{} l l rr r @{}}
\toprule
Dataset & Type & \#Train seq. & \#Eval seq. & Hours \\
\midrule
MotionMillion + LINGO & Motion--text & 913,308$^\dagger$ & 3,945$^\ddagger$ & 2,180.2$^\dagger$ \\
\midrule
HiPhi & Human--object & 13,037$^\dagger$ & 500 & 236.8$^\dagger$ \\
OMOMO & Human--object & 5,279 & 500 & 9.6 \\
HUMOTO & Human--object & 685 & 50 & 2.2 \\
\midrule
DIP-IMU & Real IMU & 41 & 19 & 1.44 \\
IMUPoser & Real IMU & 133 & 34 & 0.96 \\
IMU4D-Real & Real IMU & 75 & 27 & 1.06 \\
\midrule
ParaHome & Re-localization & 166 & 41 & 8.1 \\
\bottomrule
\end{tabular}
\end{table}

\subsection{Motion--Text Datasets}
\label{sec:supp_data_motion}
We use MotionMillion~\citep{fan2025go} as our primary data source for IMU-to-MotionText pretraining. This dataset aggregates large-scale motion sequences reconstructed from web videos and existing motion datasets. We additionally incorporate LINGO~\citep{jiang2024autonomous}, which provides 20k high-quality motion-captured sequences. We follow the original train/val/test split for MotionMillion and apply an 80/10/10 split for LINGO.

\subsection{Human--Object Interaction Datasets}
\label{sec:supp_data_hoi}

We adopt three human-object interaction datasets for IMU-to-MotionTextObject finetuning.

\noindent{\textbf{HUMOTO}}~\citep{lu2025humoto}
 is a high-fidelity human-object interaction dataset with 735 sequences (7,875 seconds at 30 fps) capturing interactions with 63 objects and 72 articulated parts. We use 685 sequences for training and 50 for testing. We use the first-frame 3D bounding box annotations of 63 object classes as ground truth, excluding articulated parts. Since HUMOTO provides motion in MIXAMO format, we apply motion retargeting to recover SMPL-X parameters.

\noindent{\textbf{HiPhi}}~\citep{ji2026hiphi}
 is a large-scale, high-precision human motion and object-interaction dataset captured with an optical motion capture system at 90 fps from 132 performers. It contains 617.5 hours of motion (including left-right mirrored sequences), comprising 371.8 hours of body-only motion and 245.7 hours of human-object interaction with synchronized object trajectories and meshes, covering 40 real-world objects from 12 categories. We use 6,519 sequences (13,037 with mirrored copies) for training and 500 for testing. Since HiPhi provides motion in BVH format, we apply motion retargeting to recover SMPL-X parameters.

\noindent{\textbf{OMOMO}}~\citep{li2023omomo}
 is a full-body human-object interaction dataset of approximately 10 hours, capturing 17 subjects manipulating 15 large everyday objects such as chairs, tables, boxes, a mop, and a vacuum. We use 5,279 sequences for training and 500 for testing. Human and object motions are recorded simultaneously with a 12-camera optical motion capture system at 120 fps, and object geometry is reconstructed from monocular videos. Since OMOMO provides SMPL-X parameters, no additional retargeting is required.

\subsection{Real IMU Datasets}
\label{sec:supp_data_real}

\noindent{\textbf{DIP-IMU}}~\citep{huang2018deep} contains real IMU recordings from 10 subjects wearing Xsens sensors, paired with reference SMPL poses. We split it by subject, using subjects 1--8 for training (41 sequences) and subjects 9--10 for testing (19 sequences). Since DIP-IMU provides SMPL poses, we retarget them to SMPL-X.

\noindent{\textbf{IMUPoser}}~\citep{mollyn2023imuposer} contains real IMU recordings from 10 participants wearing a smartphone, a smartwatch, and earbuds. We split it by participant, using participants 1--8 for training (133 sequences) and participants 9--10 for testing (34 sequences). Since IMUPoser provides SMPL poses, we retarget them to SMPL-X.

\begin{figure}[t]
    \centering
    \includegraphics[width=0.5\linewidth]{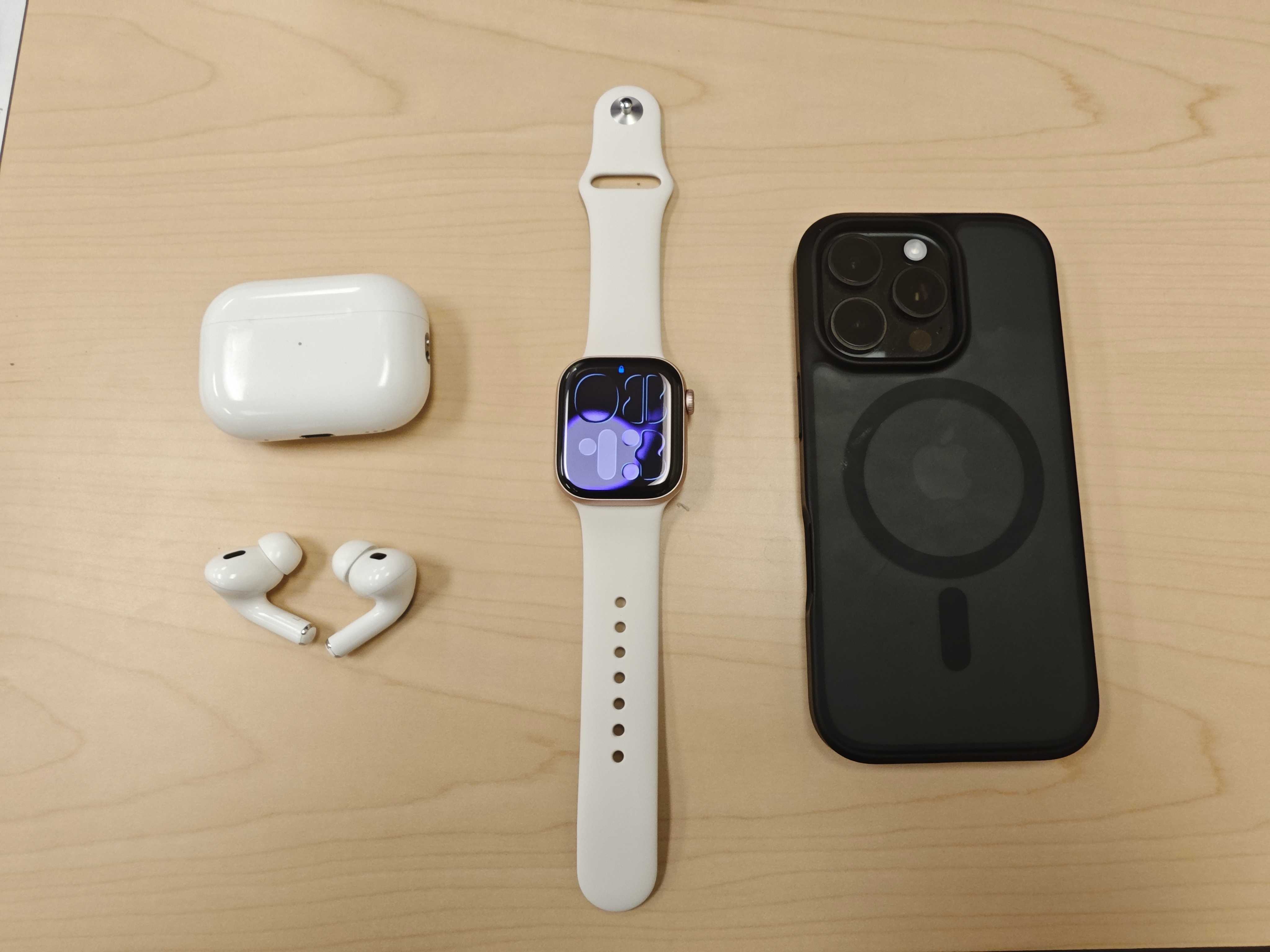}
    \caption{IMU collection device for IMU4D-Real.}
    \label{fig:device}
\end{figure}

\noindent{\textbf{IMU4D-Real}}
contains 48 recording sessions from two subjects, totaling 63.9 minutes at 30\,fps, of which 55.5 minutes are used for training and 8.4 minutes for evaluation; training and evaluation data are separated by session. As shown in \ref{fig:device}, we wear the devices on the body, including AirPods on both ears, Apple Watches on the left wrist, and iPhones in the right front pocket for all captured sequences. We use the SenorLogger app for IMU logging and synchronization. Before the motions begin, we perform a T-pose calibration for 2-3 seconds to align the IMU global frames with the coordinate conventions used during training. To obtain the ground-truth motions for training, we set up multiple RGB camera with calibrations, we predict the body motion of each frame from each camera using GVHMR~\cite{shen2024world}, and perform post-processing to merge predictions from multiple views to get a stable GT motions. And we annotate motion captions in HumanML-style for every test sequences based on the synchronized RGB videos using Qwen3-VL-32B~\citep{qwen3vl2025}. 

\subsection{Scene Re-localization Dataset}
\label{sec:supp_data_reloc}
\noindent{\textbf{ParaHome}}~\citep{kim2025parahome}
 comprises 207 high-quality SMPL-X motion sequences (486 minutes total) from 38 participants, captured with 70 synchronized RGB cameras and wearable IMU-based motion capture suits. The dataset captures full-body and hand motion with multiple objects in home environments, paired with text descriptions. We use 166 sequences for training and 41 for testing. A key characteristic of ParaHome is that the major scene layout remains consistent across sequences while human activities and interactions vary. This property makes it particularly suitable for our 3D scene re-localization task, where the goal is to situate predicted motions within a fixed environment. Therefore, we do not utilize the 3D object annotations and instead focus on localizing human motion relative to the scene layout.

\section{Quantitative Evaluation Details}
\label{sec:supp_evaluation_details}

\subsection{Metrics Definitions}

\noindent{\textbf{Motion metrics.}}
We evaluate motion quality using five metrics adopted from MobilePoser~\citep{xu2024mobileposer}. MPJPE (Mean Per Joint Position Error) measures the average Euclidean distance between predicted and ground truth joint positions. PA-MPJPE (Procrustes-Aligned MPJPE) first aligns the predicted joints to the ground truth using rigid transformation then computes MPJPE. MPJRE (Mean Per Joint Rotation Error) computes the angular differences between predicted and ground truth joint rotations. MPJVE (Mean Per Joint Vertices Error) measures the difference in human meshes between predictions and ground truth. MTE (Mean Trajectory Error) evaluates the accuracy of the root joint's global trajectory.

\noindent{\textbf{Text metrics.}}
Following MotionGPT3~\citep{zhu2025motiongpt3}, we adopt five widely-used linguistic metrics to evaluate text description quality. BERTScore~\citep{zhang2019bertscore} computes semantic similarity between predicted and reference texts using contextual embeddings from pre-trained BERT~\citep{devlin2019bert} models. BLEU@1 and BLEU@4~\citep{papineni2002bleu} measure n-gram precision with brevity penalty, where BLEU@1 focuses on unigram overlap (word-level accuracy) and BLEU@4 captures longer phrase structures up to 4-grams. ROUGE-L~\citep{lin2004rouge} computes the longest common subsequence between predicted and reference texts, emphasizing sequential word order and sentence structure. CIDEr~\citep{vedantam2015cider} weighs n-gram matches by their informativeness using TF-IDF, prioritizing distinctive and descriptive terms over common words.

\noindent{\textbf{Static object metrics.}}
For 3D object prediction, we evaluate both geometric accuracy and object recognition. We report 3D bounding box IoU (3D-IoU), which measures the volumetric overlap between predicted and ground-truth object bounding boxes. To evaluate object category recognition independently of localization quality, we additionally report identity-level precision and recall (ID-P, ID-R), which measure whether the predicted object classes match the ground-truth classes regardless of spatial accuracy. Together, these metrics capture both how well the model identifies which objects are present and how accurately it localizes them in 3D space.

\noindent{\textbf{Dynamic object metrics.}}
We evaluate dynamic objects on OMOMO and HiPhi. A frame counts as moving when the object's linear speed exceeds 0.03 m/s or its angular speed exceeds 10°/s, smoothed with a 5-frame majority filter, and a ground-truth object is dynamic if it moves in at least one frame. Predicted objects are matched one-to-one to ground-truth objects of the same category by Hungarian matching that maximizes the total first-frame oriented 3D box IoU. \emph{Recall} is the fraction of dynamic ground-truth objects that are matched, without an IoU threshold. \emph{Moving F1@0.25} is the F1 score over moving objects only, \ie, dynamic ground-truth objects and predicted objects whose predicted moving state is active in at least one frame; a matched pair counts as a true positive if its IoU is at least 0.25, and precision, recall, and F1 are computed from counts accumulated over all test sequences. \emph{Velocity error} (mm/s) is computed on matched pairs as the mean L2 difference between predicted and ground-truth per-point mesh velocities, obtained by finite differences at 30 fps over consecutive moving frames, and is averaged over the frames of a track and then over tracks.

\subsection{Evaluation Details}
For text evaluation, when multiple ground-truth captions are available for a single motion sequence, we compute the score between the predicted text and each reference caption individually, and take the maximum score as the final metric. Since the source datasets annotate captions in different styles, we rewrite the reference captions of MotionMillion, HUMOTO, OMOMO, and HiPhi into a unified style with Qwen3-0.6B; the rewriting only rephrases existing annotations, and all methods are trained and evaluated with the same rewritten captions. Both the baseline and our model are finetuned separately on each dataset before evaluation on the respective test set.

\section{Failure Cases and Limitations}
\label{sec:supp_failure}

In this section, we discuss the failure cases of our approach.

\noindent{\textbf{Drifting.}} Human motion may drift over time due to the accumulation of prediction errors. This issue is shared by all methods. We hope that incorporating loop-closure mechanisms could alleviate this problem. We leave it for future work.

\noindent{\textbf{Interpenetration.}} This occurs because our model does not explicitly enforce penetration constraints. Incorporating such penalties could help alleviate this issue.

\end{document}